%% file: main.tex
\documentclass{article}
\usepackage{amsmath,epsfig,bm, amssymb,subfigure,graphicx,algorithm,algorithmic,color}
\input{mathdef}

\usepackage{url,multirow}

\advance\oddsidemargin-1.0in
\date{\today}
\title{Localized Lasso for High-Dimensional Regression} %\thanks{Grants or other notes}

\author{Makoto Yamada$^{1}$, Koh Takeuchi$^2$, Tomoharu Iwata$^2$, John Shawe-Taylor$^3$, Samuel Kaski$^{1,4}$\\
$^1$Kyoto University, Japan\\
$^2$NTT Communication Science Laboratories, Japan\\
$^3$University College London, UK \\ 
$^4$Aalto University, Finland\\
\texttt{myamada@kuicr.kyoto-u.ac.jp, \{takeuchi.koh,iwata.tomoharu\}@lab.ntt.co.jp }\\
\texttt{j.shawe-taylor@ucl.ac.uk, samuel.kaski@aalto.fi}
}%

\begin{document}
\maketitle

\begin{abstract}
We introduce the localized Lasso, which is suited for learning models that both are interpretable and have a high predictive power in problems with high dimensionality $d$ and small sample size $n$.  More specifically, we consider a function defined by local sparse models, one at each data point. We introduce sample-wise network regularization to borrow strength across the models, and sample-wise exclusive group sparsity (a.k.a., $\ell_{1,2}$ norm) to introduce diversity into the choice of feature sets in the local models. The local models are interpretable in terms of similarity of their sparsity patterns. The cost function is convex, and thus has a globally optimal solution. Moreover, we propose a simple yet efficient iterative least-squares based optimization procedure for the localized Lasso, which does not need a tuning parameter, and is guaranteed to converge to a globally optimal solution. The solution is empirically shown to outperform alternatives for both simulated and genomic personalized medicine data.
\end{abstract}

\section{Introduction} 
A common problem in molecular medicine, shared by many other fields, is to learn predictions from data consisting of a large number of features (e.g., genes) and a small number of samples (e.g., drugs or patients). A key challenge is to tailor or ``personalize'' the predictions for each data sample. essentially solving a multi-task learning problem \cite{evgeniou2007multi,argyriou2008convex} where in each task $n=1$. The features (genes) important for predictions can be different for different samples (patients or drugs), and reporting the important features is a key part of the data analysis, requiring models that are interpretable in addition to having high prediction accuracy. That is, the problem can be regarded as a \emph{local} feature selection and prediction problem, which would be hard for existing multi-task learning approaches \cite{evgeniou2007multi,argyriou2008convex}. 

Sparsity-based linear feature selection methods such as Lasso \cite{JRSSB:Tibshirani:1996} are popular and useful for large $p$, small $n$ problems. Standard feature selection methods select the same small set of features for all samples, which is too restrictive for the multi-task type of problems, where for instance effects of different drugs may be based on different features, and dimensionality needs to be minimized due to the very small sample size.

Recently, the network Lasso \cite{hallac2015network} method has been proposed for learning local functions $f(\boldx_i;\boldw_i), i = 1,\ldots, n$, by using network (graph) information between samples. In network Lasso, a group regularizer is introduced to the difference of the coefficient vectors between linked coefficients (i.e., $\boldw_i - \boldw_j$), making them similar. We can use this regularizer to make the local models borrow strength from linked models. In the network Lasso, sparsity has so far been used only for making the coefficient vectors similar instead of for feature selection, resulting in dense models.
  
We propose a sparse variant of the network Lasso, called the localized Lasso, which helps to choose interpretable features for each sample. More specifically, we propose to incorporate the sample-wise exclusive regularizer into the network Lasso framework. By imposing the network regularizer, we can borrow strength between samples neighboring in the graph, up to clustering or ``stratifying'' the samples according to how the predictions are made. Furthermore, by imposing a sample-wise exclusive group regularizer, each learned model is made sparse but the support remains non-empty, in contrast to what could happen with naive regularization. As a result, the sparsity pattern and the weights become similar for neighboring models. We propose an efficient iterative least squares algorithm and show that the algorithm will obtain a globally optimal solution. Through experiments on synthetic and real-world datasets, we show that the proposed localized Lasso outperforms  state-of-the-art methods even with a smaller number of features.

\noindent {\bf Contribution:} 
\begin{itemize}
\item {We propose a \emph{convex} local feature selection and prediction method. Specifically, we combine the exclusive regularizer and network regularizer to produce a locally defined model that gives accurate and interpretable predictions.}
\item We propose an efficient iterative least squares based optimization procedure, which does not need a tuning parameter and is guaranteed to converge to a globally optimal solution.
\item We propose a sparse convex clustering method based on the proposed regularization.
%\item We show that the proposed method outperforms state-of-the-art network Lasso in both prediction and clustering performance.
\end{itemize}

\section{Proposed method}
In this section, we first formulate the problem and then introduce the localized Lasso.
\subsection{Problem Formulation}
Let us denote an input vector by $\boldx = [x^{(1)}, \dots, x^{(d)}]^\top \in \mathbbR^d$ and  the corresponding output value $y \in \mathbbR$. The set of samples $\{(\boldx_i, y_i)\}_{i = 1}^{\ntr}$ has been drawn i.i.d. from a joint probability density $p(\boldx, y)$. We further assume a graph $\boldR \in \mathbbR^{n \times n}$, where $[\boldR]_{i,j} = r_{ij} \geq 0$ is the coefficient that represents the relatedness between the sample pair $(\boldx_i, y_i)$ and $(\boldx_j, y_j)$. In this paper, we assume that $\boldR = \boldR^\top$ and the diagonal elements of $\boldR$ are zero (i.e., $r_{11} = r_{22} = \ldots =  r_{nn} = 0$).

{The goal in this paper is to select multiple sets of features such that each set of features is locally associated with an individual data point or a cluster, from the training input-output samples and the graph information $\boldR$. }
%The goal in this paper is to estimate a regression function, from the training input-output samples and the graph information $\boldR$. %, the function $f(\boldx)$ such that
%\begin{align*}
%y = f(\boldx) + e,
%\end{align*}
%where $e$ is noise. 
 In particular, we aim to learn a model with an interpretable sparsity pattern in the features.

\subsection{Model}
We employ the following linear model for each sample $i$:
\begin{align}
\label{eq:linmodel}
y_i = \boldw_i^\top \boldx_i.
\end{align}
Here $\boldw_i \in \mathbbR^{d}$ contains the regression coefficients for sample $\boldx_i$ and $^\top$ denotes the  transpose. Note that in regression problems  the weight vectors are typically assumed to be equal, $\boldw = \boldw_1 = \ldots = \boldw_n$. Since we cannot assume the models to be based on the same features, and we want to interpret the support of the model for each sample, we use local models. 

Since there are as many unknown variables as observed variables in Eq. \eqref{eq:linmodel}, we need to regularize, for which we propose to use network Lasso type of regularization \cite{hallac2015network}: %(first term in the following equation): %\ref{eq:struct_reg}):
\begin{align*}
%\label{eq:struct_reg}
\rho(\boldW; \boldR,\lambda_1,\lambda_2) \!=\! \lambda_1 \!\sum_{i,j = 1}^{n} r_{ij} \|\boldw_i - \boldw_j\|_{2} \!+\! \lambda_2 \!\sum_{i = 1}^{\ntr}\|\boldw_i\|_1^2.
%\lambda_1 \|\boldW\|_{2,1} + \lambda_2 \|\boldW^\top\|_{1,2}.
\end{align*}
Here $\lambda_1 \geq 0$ and $\lambda_2 \geq 0$ are the regularization parameters.  By imposing the network regularization, we regularize the model parameters $\boldw_i$ and $\boldw_j$ to be similar if $r_{ij} > 0$. If $\lambda_1$ is large, we will effectively cluster the samples according to how similar the  $\boldw_i$s are, that is, according to the prediction criteria in the local models. More specifically, when $\|\boldw_i - \boldw_j\|_2$ is small (possibly zero), we can regard the $i$-th sample and $j$-th sample to belong to the same cluster. 

The second regularization term is the $\ell_{1,2}$ regularizer (a.k.a., exclusive regularizer) \cite{kowalski2009sparse,zhou2010exclusive,kong2014exclusive}. By imposing the $\ell_{1,2}$ regularizer, we can select a small number of elements within each $\boldw_i$. Note that, we make each parameter vector $\boldw_i$ as a group (in total $n$ groups), and are not treating each dimension as a group.  Thanks to the $\ell_2$ norm over the weight vectors, the $\boldw_i$ remain non-zero (i.e., $\boldw_i \neq \boldzero$). Similarities and differences in the sparsity patterns of the $\boldw_i$ are then easily interpretable, more easily than in dense vectors. Note that while simply imposing the $\ell_1$ regularizer for all weights would induce sparsity too, for the heavy regularization required due to the small sample size, many of the $\boldw_i$ would be shrunk to zero. See Figure \ref{fig:toy} for an example.

Our proposed regularizer can be seen as a (non-trivial) extension of network regularization \cite{hallac2015network}, and hence it could be solved by a general alternating direction method of multipliers (ADMM) based solver. However, ADMM requires a tuning parameter for convergence \cite{nishihara2015general}. In this paper, we propose a simple yet effective iterative least-squares based optimization procedure, which does not need any tuning parameters, and is guaranteed to converge to a globally  optimal solution.

\begin{algorithm*}[t]
\caption{Iterative Least-Squares Algorithm for solving Eq. \eqref{eq:mkl-opt}}
\label{alg:alg}
\begin{algorithmic}
\STATE Input: $\boldZ \in \mathbbR^{n \times (dn)}$, $\boldy \in \mathbbR^n$, $\boldR \in \mathbbR^{n \times n}$, $\lambda_1$, and $\lambda_2$.
\STATE Output: $\boldW \in \mathbbR^{n\times d}$.
\STATE Set $t = 0$, Initialize $\boldF_g^{(0)}$, $\boldF_e^{(0)}$.%, $\boldU = [~]$
\REPEAT
\STATE Compute $\text{vec}({\boldW}^{(t+1)}) = (\lambda_1\boldF_g^{(t)} + \lambda_2 \boldF_e^{(t)})^{-1}\boldZ^\top (\boldI_{n} + \boldZ (\lambda_1\boldF_g^{(t)} + \lambda_2 \boldF_e^{(t)})^{-1} \boldZ^\top)^{-1}\boldy,$
\STATE Update $\boldF_g^{(t+1)}$, where $\boldF_g^{(t+1)} = \boldI_d \otimes \boldC^{(t+1)}$.
\STATE Update $\boldF_e^{(t+1)}$, where  $[\boldF_e^{(t+1)}]_{\ell,\ell} = \sum_{k = 1}^{\ntr} \frac{{I}_{k,\ell}\|{\boldw}_{k}^{(t+1)}\|_1}{[\text{vec}(|\boldW^{(t+1)}|)]_\ell}$.
\STATE $t = t + 1$.
\UNTIL{Converges}
%\RETURN $\boldW^{(T)}$.%, $\boldU$
\end{algorithmic}
\end{algorithm*}

\subsection{Optimization problem}
The optimization problem can be written as
\begin{align}
\label{eq:mkl-opt}
%\begin{split}
\min_{\boldW} \hspace{0.1cm}  J(\boldW) &= \sum_{i = 1}^n (y_i - \boldw_i^\top \boldx_i)^2 + \rho(\boldW; \boldR,\lambda_1,\lambda_2),%\lambda_1 \sum_{i,j = 1}^{n} r_{ij} \|\boldw_i - \boldw_j\|_{2} \nonumber \\
%\phantom{\min_{\boldW} \hspace{0.1cm}  J(\boldW)} &\phantom{=}+ \lambda_2 \sum_{i = 1}^{\ntr}\|\boldw_i\|_1^2, %\nonumber \\
%\end{split}
\end{align}
which is convex and hence has a globally optimal solution. Note that for classification problems the squared loss can be replaced by the logistic loss.

Let us denote $\boldX = [\boldx_1, \ldots, \boldx_n] = [\boldu_1, \ldots, \boldu_d]^\top$,  $\boldu_i \in \mathbbR^{\ntr}$, and $\boldW = [\boldw_1, \ldots, \boldw_n]^\top \in \mathbbR^{\ntr \times d}$. We can alternatively write the objective function as
\begin{align}
\label{eq:objective_function2}
%\begin{split}
{J}(\boldW) &= \|\boldy - \boldZ \text{vec}(\boldW)\|_2^2 +  \rho(\boldW; \boldR,\lambda_1,\lambda_2),%\lambda_1 \sum_{i,j = 1}^{n} r_{ij} \|\boldw_i - \boldw_j\|_{2} \nonumber \\
%&\phantom{=}+ \lambda_2 \sum_{i = 1}^{\ntr}\|\boldw_i\|_1^2,
%\end{split}
\end{align}
where  $\boldZ = \left[\textnormal{diag}(\boldu_1) ~|~ \textnormal{diag}(\boldu_2) ~|~ \ldots ~|~ \textnormal{diag}(\boldu_d) \right] \in \mathbbR^{\ntr \times ( d \ntr) }$, $\diag{\boldu} \in \mathbbR^{\ntr \times \ntr}$ is the diagonal matrix whose diagonal elements are the $\boldu$, and $\text{vec}(\cdot)$ is the vectorization operator such that
\begin{align*}
\textnormal{vec}(\boldW) &=  \Large{\textnormal{[}} [\boldW]_{1,1}, [\boldW]_{2,1}, \ldots [\boldW]_{n,1}, \ldots, \\
&\phantom{=} \hspace{0.3cm} [\boldW]_{1,d}, [\boldW]_{2,d}, \ldots [\boldW]_{n,d}\Large{\textnormal{]}}^\top \in \mathbbR^{dn}.
\end{align*}
Here we use the vectorization operator since it makes it possible to write the loss function and the two regularization terms as a function of $\textnormal{vec}(\boldW)$, which is highly helpful for deriving a simple update formula for $\boldW$.

Taking the derivative of $J(\boldW)$ with respect to $\textnormal{vec}(\boldW)$ and using the Propositions \ref{prop_derive_net} and \ref{prop_derive_exc} (See Appendix), the optimal solution is given as
\begin{align}
\label{eq:optimum_solution}
\textnormal{vec}(\boldW) = ( \boldZ^\top  \boldZ  + \lambda_1 \boldF_g  + \lambda_2 \boldF_e)^{-1} \boldZ^\top \boldy,
\end{align}
where
\begin{align*}
\boldF_g &= \boldI_d \otimes \boldC,~[\boldF_e]_{\ell,\ell} = \sum_{i = 1}^{\ntr} \frac{I_{i,\ell}\|\boldw_{i}\|_1}{[\text{vec}(|\boldW|)]_\ell},\\
[\boldC]_{i,j} &=\left\{ \begin{array}{ll}
\sum_{j' = 1}^n \frac{r_{ij'}}{\|\boldw_{i} - \boldw_{j'}\|_{2}} -\frac{r_{ij}}{\|\boldw_i - \boldw_j\|_{2}} & (i = j) \\
\frac{-r_{ij}}{\|\boldw_i - \boldw_j\|_{2}} & (i \neq j) 
\end{array} \right. .
\end{align*}
Here $\boldF_e$ is diagonal, $\boldI_d \in \mathbbR^{d\times d}$ is the identity matrix, $\otimes$ is the Kronecker product, and the $I_{i,\ell} \in \{0,~1\}$ are group index indicators: $I_{i,\ell} = 1$ if the $\ell$-th element $[\text{vec}(\boldW)]_{\ell}$ belongs to group $i$ (i.e., $[\text{vec}(\boldW)]_{\ell}$ is the element of $\boldw_i$), otherwise $I_{i,\ell} = 0$. 

Since the optimization problem Eq.~\eqref{eq:mkl-opt} is convex, the $\boldW$ is a global optimum to the problem if and only if  Eq.~\eqref{eq:optimum_solution} is satisfied. However, the matrices $\boldF_g$ and $\boldF_e$ are dependent on $\boldW$ and are also unknown. Thus, we instead optimize the following objective function to solve Eq.~\eqref{eq:mkl-opt}:
\begin{align}
\label{eq:objective_function3}
\begin{split}
\widetilde{J}(\boldW) &= \|\boldy - \boldZ \text{vec}(\boldW)\|_2^2 \\
&\phantom{=} + \text{vec}(\boldW)^\top (\lambda_1 \boldF_g^{(t)}  + \lambda_2 \boldF_e^{(t)})\text{vec}(\boldW),
\end{split}
\end{align}
where $\boldF_g^{(t)} \in \mathbbR^{d\ntr \times d\ntr}$ is a block diagonal matrix and $\boldF_e^{(t)} \in \mathbbR^{d\ntr \times d\ntr}$ is a diagonal matrix whose diagonal elements are defined as\footnote{When $\boldw_i - \boldw_j = \boldzero$, then $\boldF_g$ is the subgradient of $\sum_{i,j=1}^{\ntr} r_{ij}\|\boldw_i - \boldw_j\|_2$. Also, $\boldF_e$ is the subgradient of $\sum_{i = 1}^n \|\boldw_i\|_{1}^2$ when $[\text{vec}(|\boldW|)\|_{\ell} = 0$. However, we cannot set elements of $\boldF_g$ to 0 (i.e.,  when $\boldw_i - \boldw_j = \boldzero$) or the element of $[\boldF_e]_{\ell,\ell} = 0$ (i.e., when $[\text{vec}(|\boldW|)\|_{\ell} = 0$), otherwise the Algorithm~1 cannot be guaranteed to converge. To deal with this issue, we can use $\sum_{i,j=1}^{\ntr} r_{ij}\|\boldw_i - \boldw_j + \epsilon \|_2$ and $\sum_{i = 1}^n \|\boldw_i + \epsilon\|_{1}^2$ ($\epsilon > 0$) instead \cite{kong2014exclusive,nie2010efficient}. }
\begin{align*}
\boldF_g^{(t)} &\!=\! \boldI_d \otimes \boldC^{(t)},~[\boldF_e]_{\ell,\ell}^{(t)} \!=\! \sum_{i = 1}^{\ntr}\! \frac{I_{i,\ell}\|\boldw_{i}^{(t)}\|_1}{[\text{vec}(|\boldW^{(t)}|)]_\ell},\\
[\boldC^{(t)}]_{i,j} &\!=\!\left\{\!\! \begin{array}{ll}
\sum_{j' = 1}^n \!\frac{r_{ij'}}{\|\boldw_{i}^{(t)} - \boldw_{j'}^{(t)}\|_{2}} \!-\! \frac{r_{ij}}{\|\boldw_i^{(t)} \!-\! \boldw_j^{(t)}\|_{2}} & (i = j) \\
\frac{-r_{ij}}{\|\boldw_i^{(t)} - \boldw_j^{(t)}\|_{2}} & (i \neq j) 
\end{array} \right. 
\end{align*}

We propose to use the iterative least squares approach to optimize Eq.~\eqref{eq:objective_function3}. With given  $\boldF_g^{(t)}$ and $\boldF_e^{(t)}$, the optimal solution of $\boldW$ is obtained by solving $\frac{\partial \widetilde{J}(\boldW)}{\partial \boldW} = \boldzero$. The $\boldW$ is estimated as %by solving the following linear equation:
%\begin{align*}
%-2 \boldA^\top (\boldy - \boldA \text{vec}(\boldW)) + 2 (\lambda_1 \boldF_g^{(t)}  + \lambda_2 \boldF_e^{(t)})\text{vec}(\boldW) &= \boldzero \\
%( \boldZ^\top \boldZ  + \lambda_1 \boldF_g^{(t)}  + \lambda_2 \boldF_e^{(t)}) \text{vec}(\boldW)  &= \boldZ^\top \boldy.
%\end{align*}
%Thus, $\boldW^{(t+1)}$ can be obtained as
%\begin{align*}
%\text{vec}(\boldW^{(t+1)}) = ( \boldZ^\top  \boldZ  + \lambda_1 \boldF_g^{(t)}  + \lambda_2 \boldF_e^{(t)})^{-1} \boldZ^\top \boldy.
%\end{align*}
%However, since $\boldZ^\top \boldZ  + \lambda_1 \boldF_g^{(t)}  + \lambda_2 \boldF_e^{(t)} \in \mathbbR^{d\ntr \times d\ntr}$ is large and can be dense due to $\boldZ^\top \boldZ$, it is hard to solve the linear equation directly. Hence, we employ the  Woodbury formula \cite{petersen2008matrix}:
%\[
%(\boldZ^\top \boldZ + \boldF)\boldZ^\top = \boldF^{-1} \boldZ^\top (\boldI_n + \boldZ \boldF^{-1} \boldZ^\top)^{-1}.
%\]
% Finally, $\text{vec}({\boldW}^{(t+1)})$ is obtained as
\begin{align}
\label{eq:update_beta}
\text{vec}({\boldW}^{(t+1)}) \!=\! {\boldH^{(t)}}^{-1}\boldZ^\top (\boldI_{n} \!+\! \boldZ {\boldH^{(t)}}^{-1} \boldZ^\top)^{-1}\boldy,
\end{align}
where $\boldH^{(t)} = \lambda_1\boldF_g^{(t)} + \lambda_2 \boldF_e^{(t)}$, $\boldF_g^{(t)}$ is block diagonal and $\boldF_e^{(t)}$ diagonal. Here, we employ the  Woodbury formula \cite{petersen2008matrix}. After we obtain $\boldW^{(t+1)}$, we update $\boldF_g^{(t+1)}$ and $\boldF_e^{(t+1)}$. We continue this two-step procedure until convergence. The algorithm is summarized in Algorithm \ref{alg:alg}.

%More specifically, we iterate the following two steps until convergence:
%\begin{itemize}
%\item  Update $\boldW^{(t+1)}$ with $\boldF_g^{(t)}$ and $\boldF_e^{(t)}$.
%\item  Update $\boldF_g^{(t+1)}$ and $\boldF_e^{(t+1)}$ with $\boldW^{(t+1)}$.
%\end{itemize}

% \vspace{.1in}
%\noindent {\bf Other regularizations:} It can easily incorporate other regularization terms into our optimization framework if the regularization term can be written as
%\[
%\text{vec}(\boldW)^\top \boldF^{(t)}\text{vec}(\boldW).
%\]
%For example, the group regularization (i.e., $\ell_{2,1}$ norm) can be written in this form \cite{nie2010efficient}, and thus, it can be included.
 
 \vspace{.1in}
\noindent {\bf Predicting for new test sample:}
For predicting on test sample $\boldx$, we use the estimated local models $\widehat{\boldw}_k$ which are linked to the input $\boldx$. More specifically, we solve the Weber problem \cite{hallac2015network}
\begin{align}
\label{eq:weber}
\min_{\boldw} \hspace{0.3cm} \sum_{i = 1}^n r'_{i} \|\boldw - \widehat{\boldw}_i\|_2,
\end{align}
where $r'_i \geq 0$ is the link information between the test sample and the training sample $\boldx_i$. Since this problem is convex, we can solve it efficiently by an iterative update formula (see Algorithm \ref{alg:weber}). If there is no link information available, we simply average all $\widehat{\boldw}_i$s to estimate $\widehat{\boldw}$, and then predict  as $\widehat{y} = \widehat{\boldw}^\top \boldx$.

\begin{algorithm}[t]
\caption{Iterative Least-Squares Algorithm for solving Eq. \eqref{eq:weber}}
\label{alg:weber} 
\begin{algorithmic}
\STATE Input: $\boldx$, $\boldr' \in \mathbbR^n$, and $\widehat{\boldW} \in \mathbbR^{d \times n}$.
\STATE Output: $\widehat{y} \in \mathbbR$ and $\widehat{\boldw} \in \mathbbR^{d}$.
\STATE Set $t = 0$, Initialize $\boldf_g \in \mathbbR^{n}$.
\REPEAT
\STATE Compute $\boldw^{(t+1)} = \frac{1}{\boldone_n^\top \boldf_g^{(t)}}\widehat{\boldW}\boldf_g^{(t)}$.
\STATE Update $\boldf_g^{(t+1)}$, where $[\boldf_g^{(t+1)}]_i = \frac{[\boldr']_i}{2\|\widehat{\boldw}_i - \boldw^{(t+1)}\|_2} $.
\STATE $t = t + 1$.
\UNTIL{Converges}
\STATE $\widehat{y} = \widehat{\boldw}^\top \boldx$.
%\RETURN $\boldW^{(T)}$.%, $\boldU$
\end{algorithmic}
\end{algorithm}

\subsection{Convergence Analysis} 
Next, we prove the convergence of the algorithm. 
\begin{theo}
\label{theo:theo1}
The Algorithm 1 will monotonically decrease the objective function Eq.~\eqref{eq:mkl-opt} in each iteration, and converge to the global optimum of the problem. 

\vspace{.1in}
\noindent Proof: Under the updating rule of Eq. \eqref{eq:update_beta}, we have the following inequality using Lemma \ref{lemm1} and Lemma \ref{lemm5} (See Appendix):
\begin{align*}
{J}(\boldW^{(t+1)}) - {J}(\boldW^{(t)}) \leq \widetilde{J}(\boldW^{(t+1)}) - \widetilde{J}(\boldW^{(t)}) \leq 0.
%J(\boldW^{(t+1)}) \leq J(\boldW^{(t)}).
\end{align*}
That is, the Algorithm 1 will monotonically decrease the objective function of Eq.~\eqref{eq:mkl-opt}. At convergence, $\boldF_g^{(t)}$ and $\boldF_e^{(t)}$ will satisfy Eq.~\eqref{eq:optimum_solution}. Since the optimization problem Eq.~\eqref{eq:mkl-opt} is convex, satisfying Eq.~\eqref{eq:optimum_solution} means that $\boldW$ is a global optimum to the problem in Eq.~\eqref{eq:mkl-opt}. Thus, the Algorithm 1 will converge to the global optimum of the problem  Eq.~\eqref{eq:mkl-opt}. \proofend 
\end{theo}

\subsection{Other applications (Sparse convex clustering)}
The proposed sparse regularization can be applied to convex clustering problems \cite{pelckmans2005convex,hocking2011clusterpath,wang2016sparse} by changing the objective function. The optimization problem is then
\begin{align}
\label{eq:opt_clustering}
\min_{\boldW} \hspace{0.1cm}  J(\boldW) &= \|\boldX^\top - \boldW\|_{F}^2 + \rho(\boldW; \boldR,\lambda_1,\lambda_2), 
%\lambda_1 \sum_{i,j = 1}^{n} r_{ij} \|\boldw_i - \boldw_j\|_{2} \nonumber \\
%&\phantom{=}+ \lambda_2 \sum_{i = 1}^{\ntr}\|\boldw_i\|_1^2, %\nonumber \\
\end{align}
where 
\[r_{ij} =\left\{ \begin{array}{ll}
\delta_{ij} \exp \left(-\frac{\|\boldx_i - \boldx_j\|^2}{2}\right) & (i \neq j) \\
0 & \textnormal{Otherwise}
\end{array} \right. .
\]
Here $\delta_{ij} = 1$ if $\boldx_j$ is included in the $K$-th neighbors of $\boldx_i$, otherwise $\delta_{ij} = 0$. The original convex clustering methods do not include the exclusive group sparsity regularization, and thus, the learned matrix $\boldW$ tends to be dense. Adding the sparsity makes the clusters more easily interpretable, even as biclusters or subspace clusters, still retaining convexity. 

%This optimization problem can similarly solved by using the iterative least squares algorithm. The updating rule of $\boldW$ is given as 
%\begin{align*}
%\text{vec}(\boldW^{(t+1)}) = ( \boldI_{dn}  + \lambda_1 \boldF_g^{(t)}  + \lambda_2 \boldF_e^{(t)})^{-1} \textnormal{vec}(\boldX^\top).
%\end{align*}
%Since the loss function is convex, we can similarly proof that the learned $\widehat{\boldW}$ is a global optimum of the optimization problem Eq. \eqref{eq:opt_clustering}.

%\subsection{Other regularization combinations}

\section{Related Work}
In this section, we review the existing regression methods and address the difference from the proposed method.

Sparsity-based global feature selection methods such as Lasso \cite{JRSSB:Tibshirani:1996} are useful for selecting genes. However, in personalized medicine setups, we ultimately want to personalize the models for each patient (or drug), instead of assuming the same set of features (e.g., genes) for each.

%In \cite{tang2012feature}, they proposed to use link (graph) information for feature selection. More specifically, they introduced a graph regularizer to enhance feature selection performance. However, similar to Lasso, the method aims to find a single set of features for all samples. Thus, it cannot find a separate set of features for each sample.

The proposed method is also related to the fused Lasso \cite{tibshirani2005sparsity}, which is widely used for analyzing spatial signals including brain signals \cite{xin2014efficient,ren2015scalable}. Both the fused Lasso and its generalizations \cite{takeuchi2015higher} operate on differences of scalars and are not suited for  the differences of vectors we would need.

The generalized group fused Lasso \cite{pelckmans2005convex,hocking2011clusterpath} is a multivariate extension of the generalized fused Lasso, used for convex clustering problems. The key difference from the original convex clustering methods and our work is the exclusive regularization term, which enables us to select features in addition to clustering samples.  Recently, a sparse convex clustering method has been proposed \cite{wang2016sparse}; its combination of feature-wise group regularization and sample-wise group fused regularization tends to select global features important for all samples, whereas we can choose features specific to each cluster and sample.

The network Lasso \cite{hallac2015network} is a general framework for solving regression problems having graph information, and our task can be categorized as a network Lasso problem. To our knowledge, ours is the first work to introduce feature-wise sparsity in the network Lasso problem. The additional central insight we bring is that by using the $\ell_{1,2}$ regularizer instead of the straightforward $\ell_1$, we get non-obvious effects resulting in learning of different sparsity patterns for each local model, still borrowing strength according to the network.

{
Multi-task learning \cite{argyriou2008convex,evgeniou2007multi,obozinski2006multi,zhou2010exclusive} is also relevant but does not solve our problem setup, since multi-task learning approaches assume the tasks (or clusters) to be known a priori. In contrast, in the localized lasso problem the clusters need to be found in addition to selecting features. It is possible to first cluster based on the similarities in $\boldR$ and then apply multi-task learning for the resulting clusters. Convex multi-task learning methods which share inter-task similarity through low-rankness exist \cite{ando2005framework,jacob2009clustered}, but have not been designed to select a small number of features for each task. Recently, FORMULA, which both shares inter-task information using low-rankness, and enforces the low-rank matrices to be sparse, has been proposed \cite{xuformula}, and we compare with it experimentally. However, FORMULA is a non-convex method, and it tends to perform poorly unless initialized very carefully. In particular in personalized medicine problems, since the data tend to be high-dimensional (i.e., the number of samples is much smaller than that of features), it tends to get trapped to poor local optima.  Since our proposed method is \emph{convex} and can effectively handle the joint feature selection and clustering problem, it is directly suited to such problem setups.} 
%The graph guided exclusive regularization \cite{chen2011towards} is seemingly related, but the important difference is that it uses \emph{feature-wise} exclusive group regularization instead of our \emph{sample-wise} regularization. As a result, all coefficients of some local models can and will be shrunk to zero, making the model non-suited to our task. Additionally, the graph regularization is $\ell_2^2$ based instead of sparse, which does not result in clustering. 

\section{Experiments}
In this section, we first illustrate our proposed method on synthetic data and then compare it with existing methods using a real-world dataset.

We compared our proposed method with Lasso \cite{JRSSB:Tibshirani:1996}, Elastic Net 
\cite{zou2005regularization}, FORMULA \cite{xuformula}, and Network Lasso \cite{hallac2015network,hallac2015snapvx}. For Lasso, Elastic Net, and FORMULA, we used the publicly available packages. For the network Lasso implementation, we set the regularization parameter to  $\lambda_2 = 0$ in the localized Lasso. For supervised regression problems, all tuning parameters are determined by 3-fold nested cross validation.

The experiments were run on a 3GHz AMD Opteron Processor with 48GB of RAM.
 
\subsection{Synthetic experiments (High-dimensional regression)}
We illustrate the behavior of the proposed method using a synthetic  high-dimensional dataset.

In this experiment, we first generated the input variables as $x_{k,i} \sim \text{Unif}(-1,1), k = 1, \ldots, 10, i = 1, \ldots, 30$. Then, we generated the corresponding output as
\begin{eqnarray}
y_{i}=\left\{ \begin{array}{ll}
5x_{1,i} + x_{2,i} - x_{3,i} + 0.1 e_i & (i = 1, \ldots, 10) \\
x_{2,i} - 5x_{3,i} + x_{4,i} + 0.1e_i & (i = 11 \ldots, 20) \\
0.5x_{4,i} - 0.5x_{5,i} + 0.1e_i & (i = 21 \ldots 30) \\
\end{array} \right., 
\end{eqnarray} 
where $x_{k,i}$ is the value of the $k$-th feature in the $i$-th sample and $e_i \sim N(0,1)$.
 In addition to the input-output pairs, we also randomly generated the link information matrix $\boldR \in \{0,1\}^{30 \times 30}$. In the link information matrix, only 40\% of true links are observed.% (See Figure \ref{fig:toy_link}(a)). 
 We experimentally set the regularization parameter for the proposed method to $\lambda_1 = 5$ and $\lambda_2 = \{0.01, 1, 10\}$. For the network Lasso, we used $\lambda_1 = 5$. Moreover, we compared the proposed method with the network Lasso + $\ell_1$ regularizer, in which we used $\lambda_1 = 5$ and $\lambda_2 = \{0.05, 0.5\}$, where $\lambda_2$ is the regularization parameter for the $\ell_1$ regularizer.
 
   \begin{figure}[t!]
\begin{center}
%\begin{minipage}[t]{0.325\linewidth}
%\centering
%  {\includegraphics[width=0.99\textwidth]{graphinfo_toy.eps}} \\ \vspace{-0.10cm}
%(a) Link information.
%\end{minipage}
\begin{minipage}[t]{0.475\linewidth}
\centering
  {\includegraphics[width=0.99\textwidth]{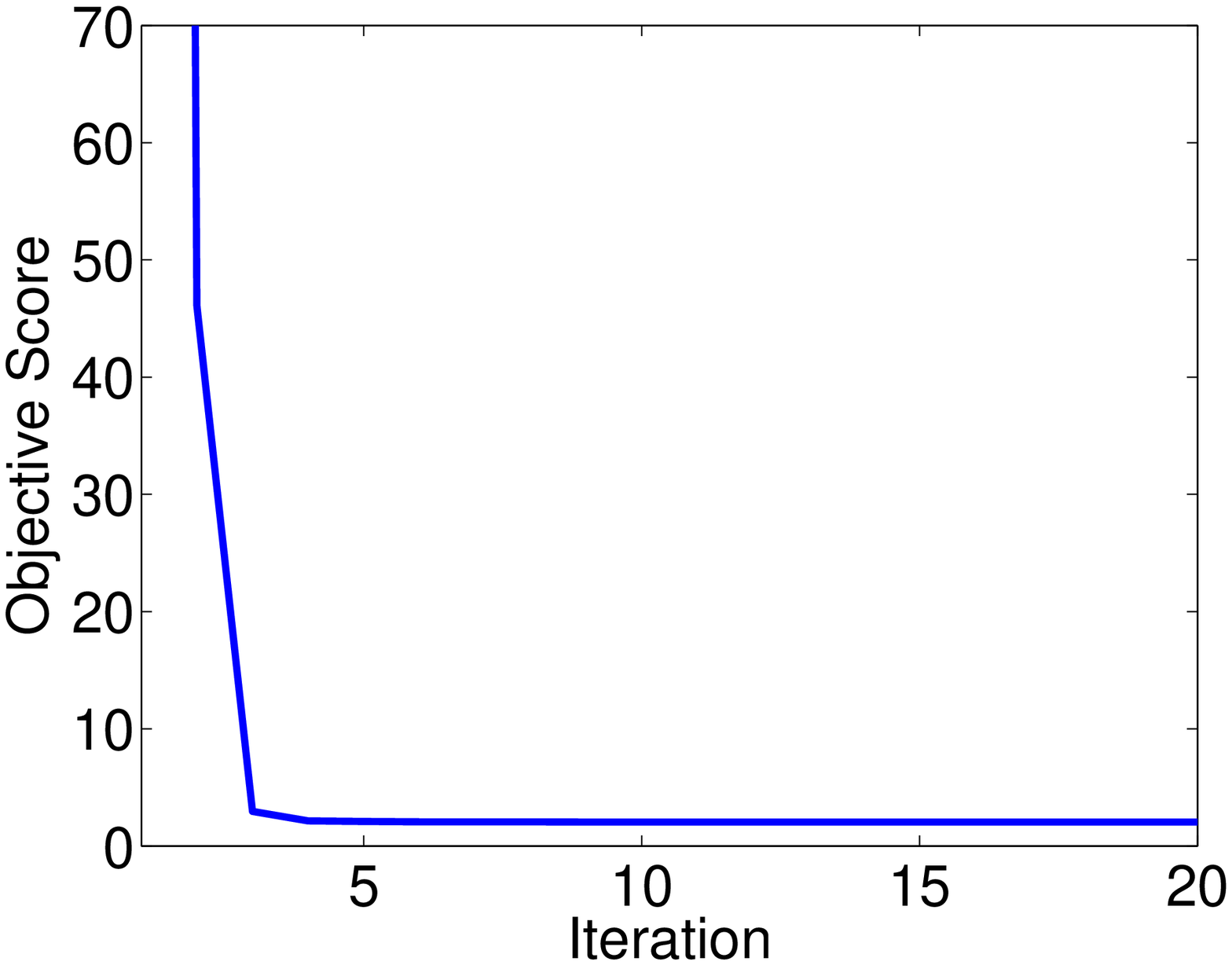}} \\ \vspace{-0.10cm}
(a) Convergence.
\end{minipage}
\begin{minipage}[t]{0.475\linewidth}
\centering
  {\includegraphics[width=0.99\textwidth]{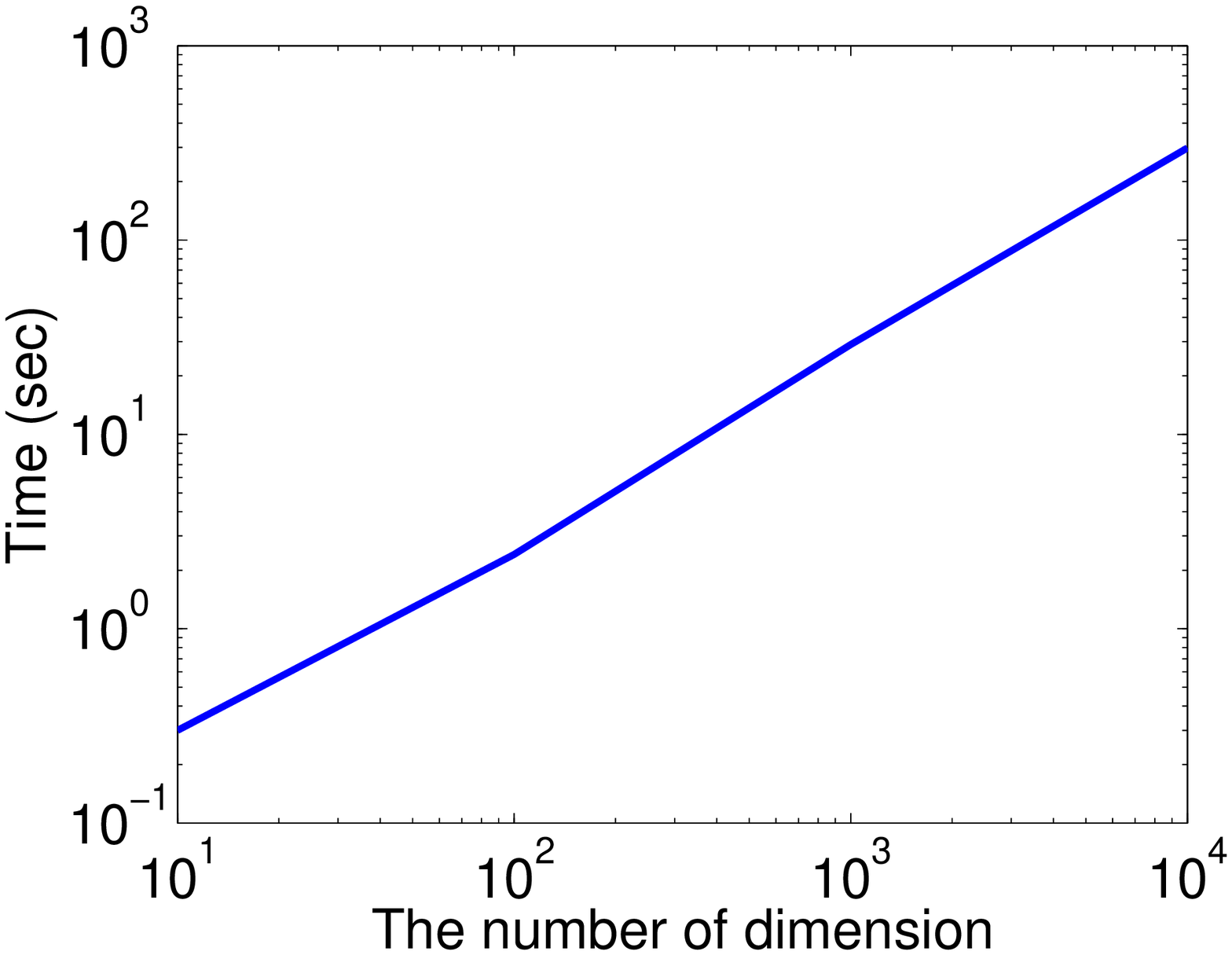}} \\ \vspace{-0.10cm}
(b) Computation time.
\end{minipage}
\caption{%(a): The link information between samples in the simulated data.
(a): Objective score (eqn~\ref{eq:mkl-opt}) are a function of iteration. (b): Computation time of the proposed method. We fixed the number of samples to 100 and the number of iterations to 10, and computed the results for  dimensions 10, 100, 1000, and 10000. }
    \label{fig:toy_link}
\end{center}
%\vspace{-.2in}
%\vspace{-.15in}
\end{figure}

  \begin{figure*}[t!]
\begin{center}
\begin{minipage}[t]{0.325\linewidth}
\centering
  {\includegraphics[width=0.99\textwidth]{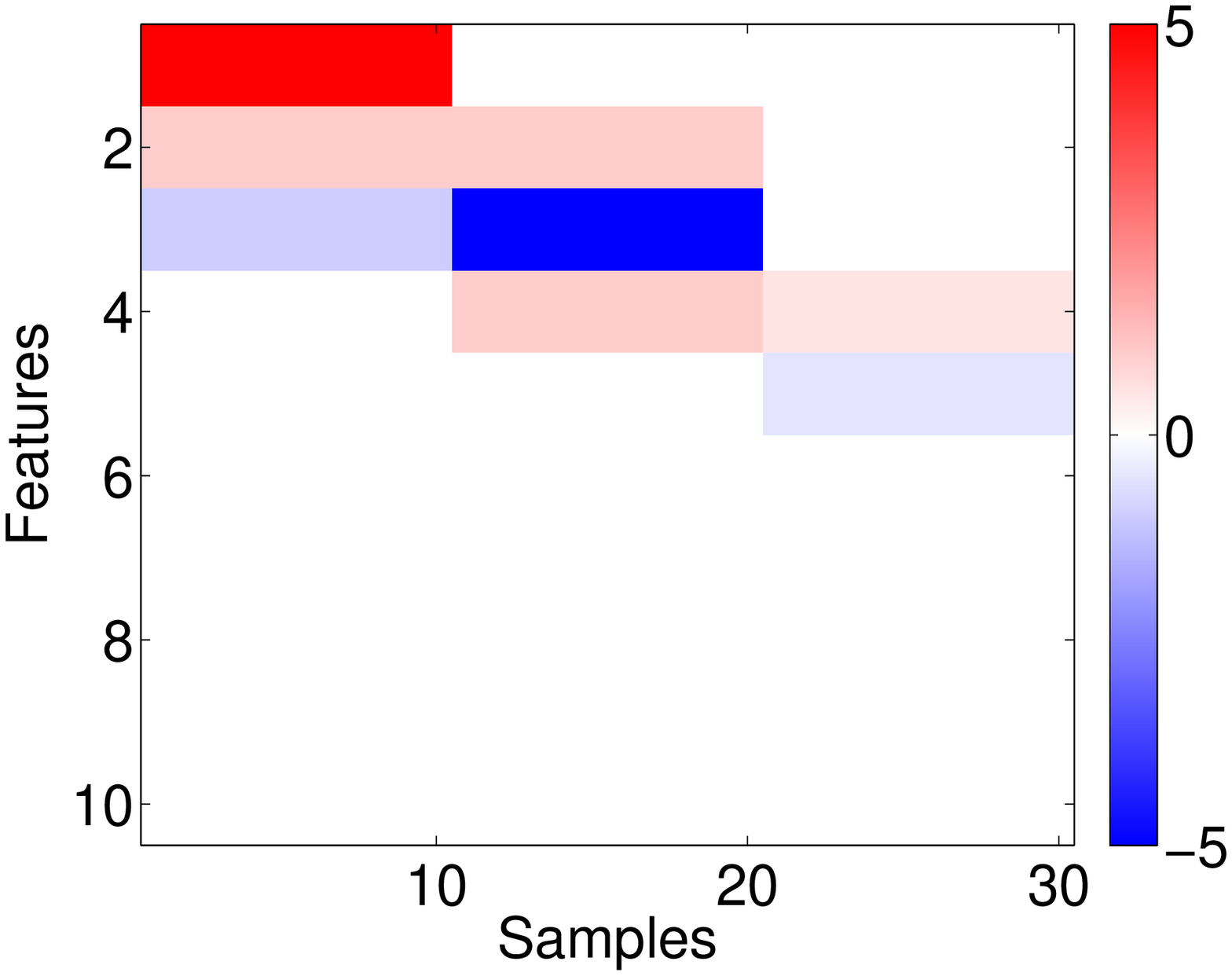}} \\ \vspace{-0.10cm}
(a) True pattern.
\end{minipage}
\begin{minipage}[t]{0.325\linewidth}
\centering
  {\includegraphics[width=0.99\textwidth]{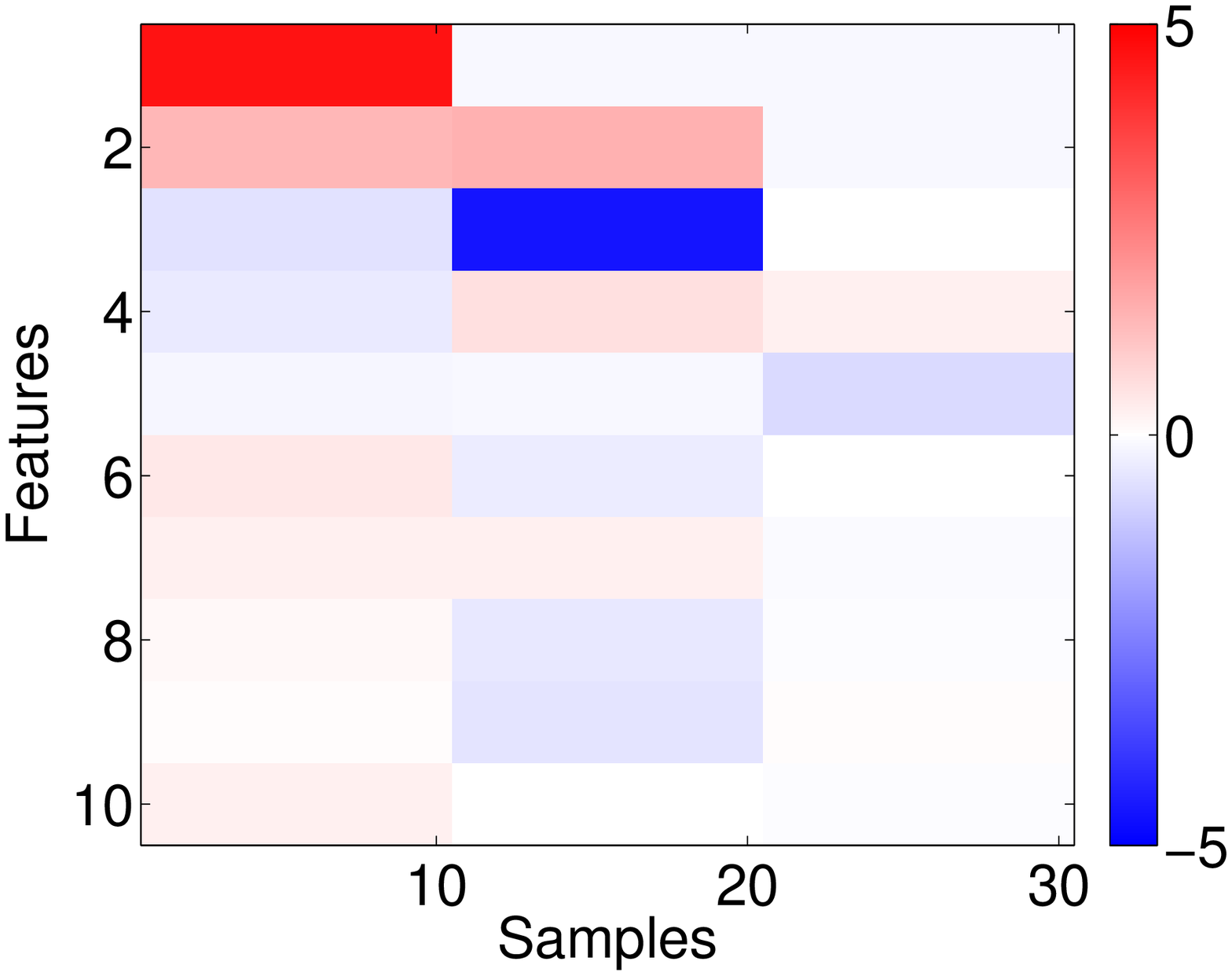}} \\ \vspace{-0.10cm}
(b) Network Lasso.
\end{minipage}
\begin{minipage}[t]{0.325\linewidth}
\centering
  {\includegraphics[width=0.99\textwidth]{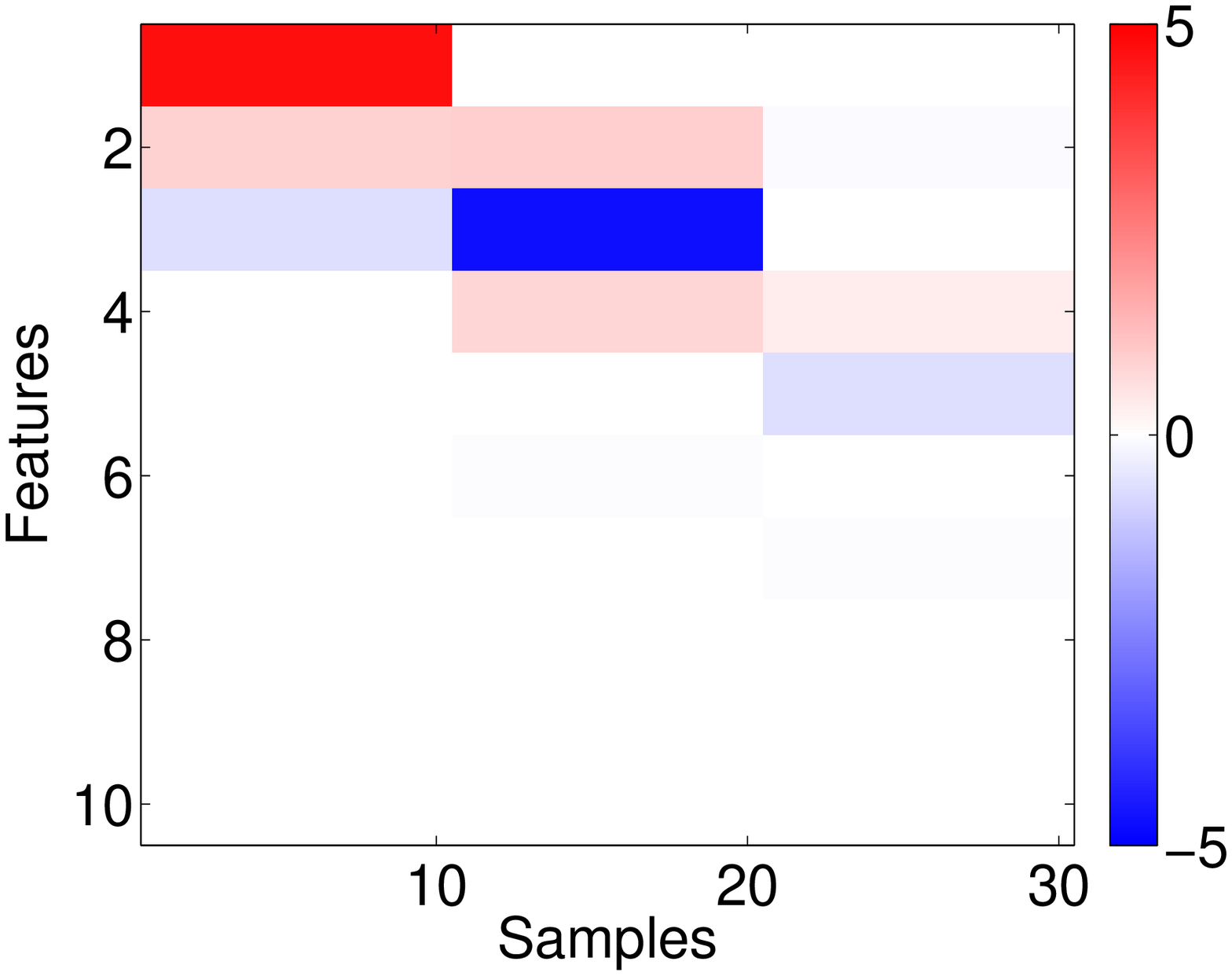}} \\ \vspace{-0.10cm}
(c) Proposed ($\lambda_2 = 0.01$).
\end{minipage}
\begin{minipage}[t]{0.325\linewidth}
\centering
  {\includegraphics[width=0.99\textwidth]{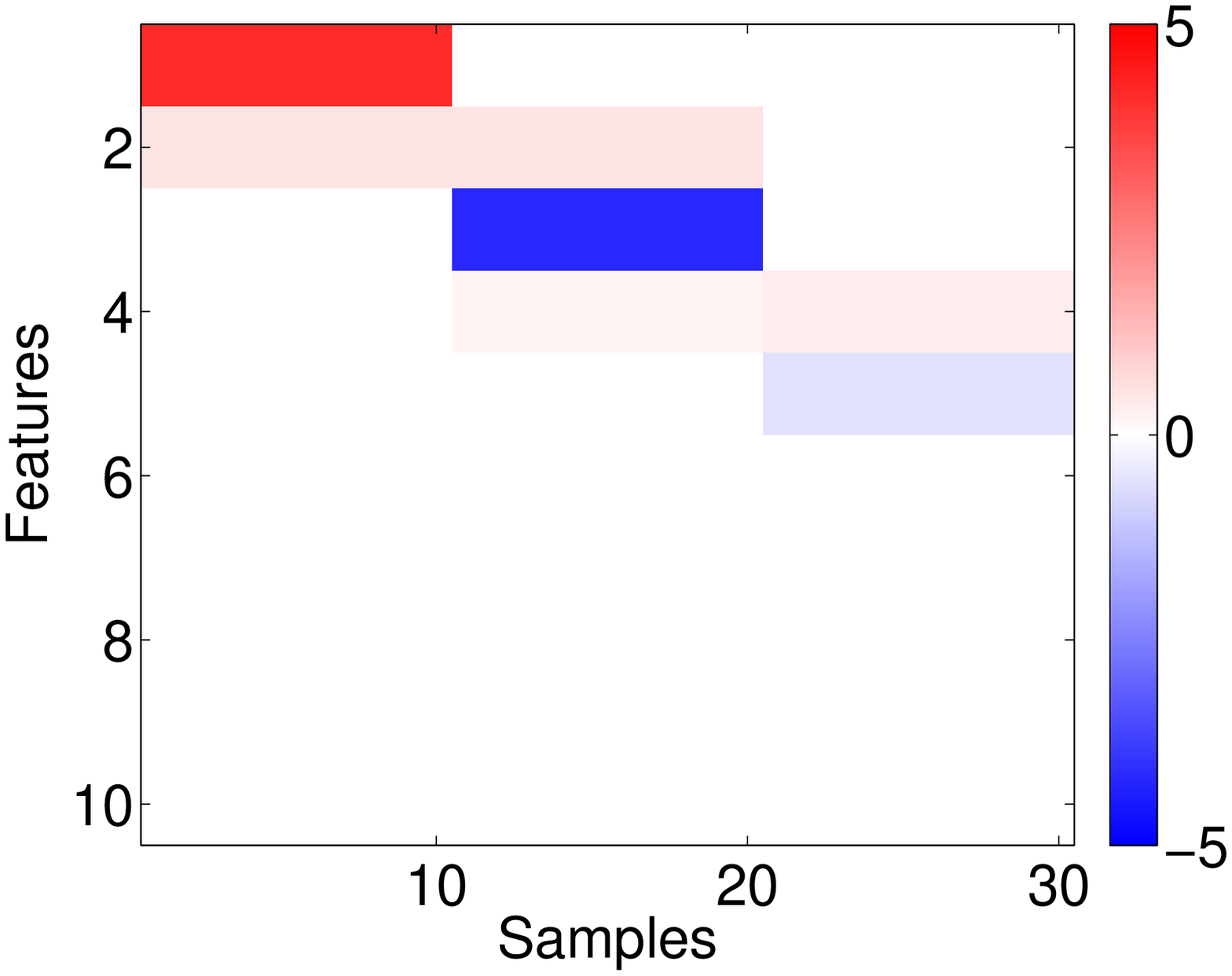}} \\ \vspace{-0.10cm}
(d) Proposed ($\lambda_2 = 0.05$).
\end{minipage}
\begin{minipage}[t]{0.325\linewidth}
\centering
 {\includegraphics[width=0.99\textwidth]{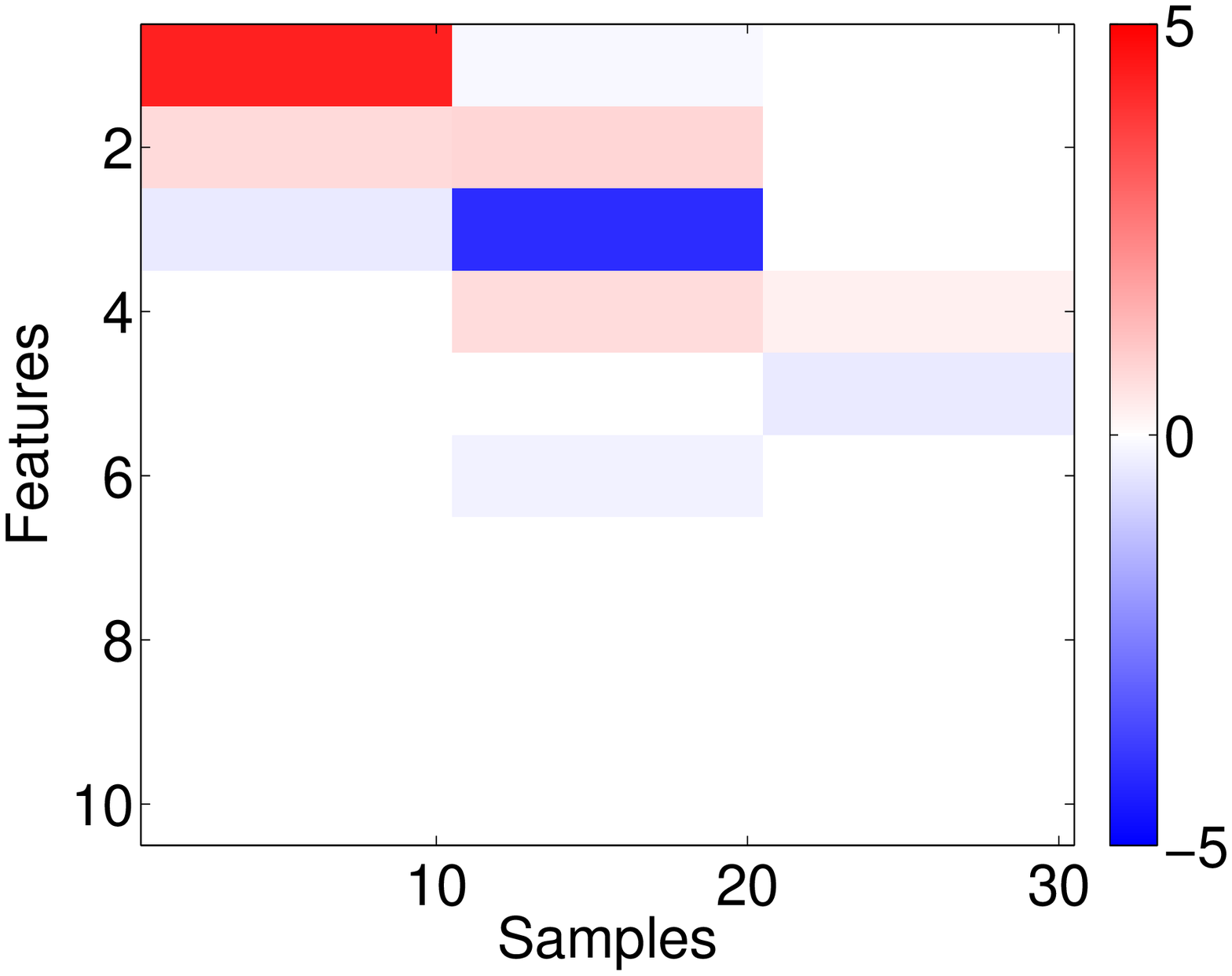}} \\ \vspace{-0.10cm}
(e) Network Lasso + $\ell_1$ ($\lambda_2 = 0.1$).
\end{minipage} 
\begin{minipage}[t]{0.325\linewidth}
\centering
 {\includegraphics[width=0.99\textwidth]{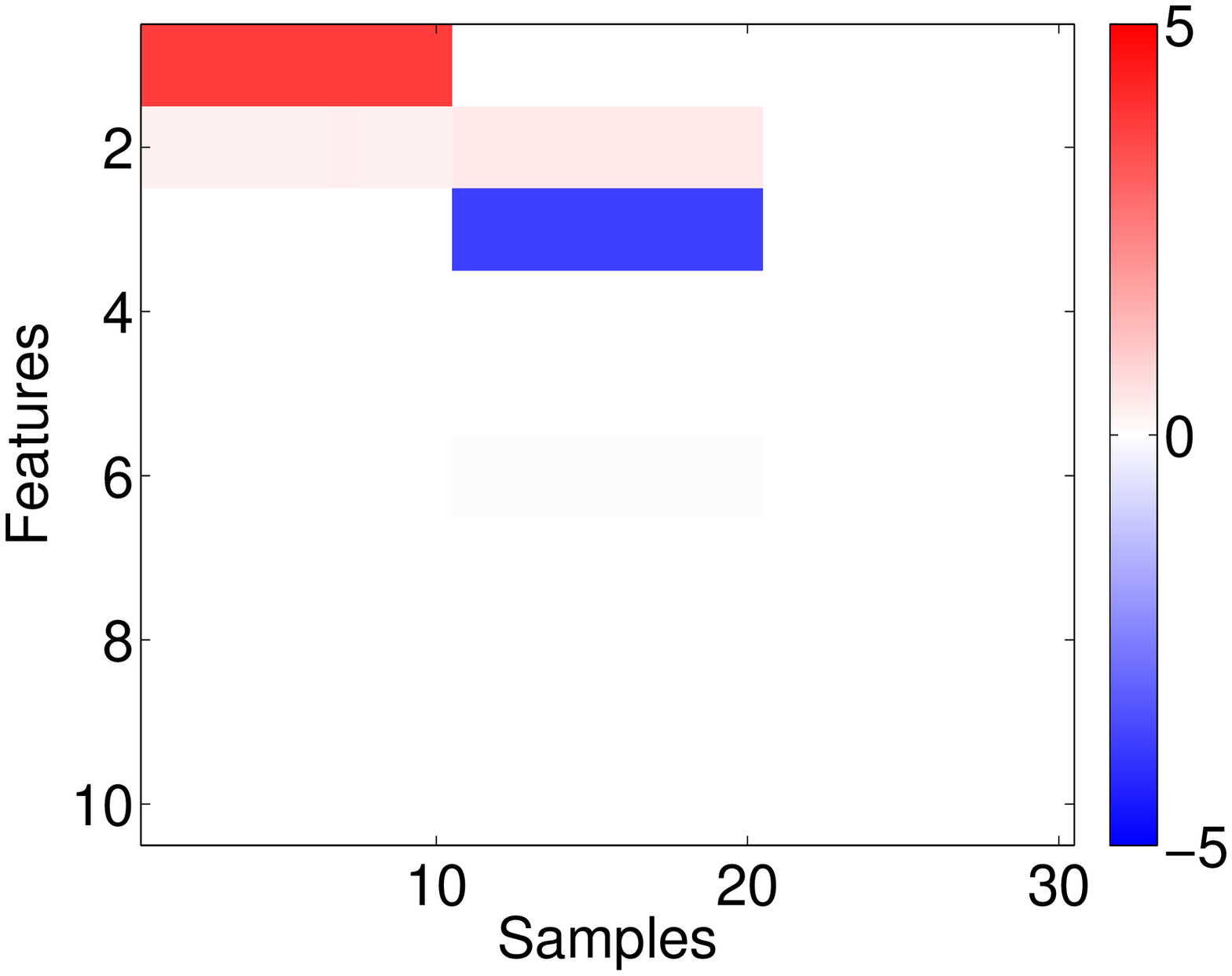}} \\ \vspace{-0.10cm}
(f) Network Lasso + $\ell_1$ ($\lambda_2 = 0.3$).
\end{minipage}
\caption{The learned coefficient matrix for the synthetic data for the different methods. Proposed $=$ Localized Lasso. For Network Lasso + $\ell_1$, we use the $\ell_1$ regularizer instead of $\ell_{1,2}$ and $\lambda_2 \geq 0$ is the regularization parameter for the $\ell_1$ term. }
    \label{fig:toy}
\end{center}
%\vspace{-.2in}
%\vspace{-.15in}
\end{figure*}
 
 Figures \ref{fig:toy}(a)-(f) show the true coefficient pattern and the results of the learned coefficient matrices $\boldW$ by using the localized Lasso, the network Lasso, and\footnote{Note that the combination of the network lasso and $\ell_1$ is also new.} the network Lasso + $\ell_1$. As can be seen, most of the unrelated coefficients of the proposed method are shrunk to zero. On the other hand, for the network Lasso, many unrelated coefficients take non-zero values. Thus, by incorporating the exclusive regularization in addition to the network regularization, we can learn sparse patterns in high-dimensional regression problems. Moreover, by setting the $\ell_{1,2}$ regularizer term to be stronger, we can obtain a sparser pattern within the $\boldw_i$. In contrast, Network Lasso + $\ell_1$, which produces a similar pattern when the regularization is weak (Fig.~\ref{fig:toy}(e)), shrinks many local models to zero if the regularization parameter $\lambda_2$ is large (Fig.~\ref{fig:toy}(f)). This shows that the network Lasso + $\ell_1$ is sensitive to the setting of the regularization parameter. Moreover, since we want to interpret features for each sample (or model), the $\ell_{1,2}$ norm is more suited than $\ell_1$ for our tasks. 
 
 Figure \ref{fig:toy_link} (a) shows the convergence of the proposed method. The objective score converges within a few iterations, without requiring tuning of step-size parameters as ADMM optimization would. Figure \ref{fig:toy_link} (b) shows the computation time of the proposed method (implemented with the same optimization package, the alternatives would perform similarly). As can be seen, the proposed method scales linearly with respect to the dimension.

\begin{table*}[t!]
{
\begin{center}
\caption{Test root MSE on toxicogenomics data. The best method under sigfinicance level 5\% (Wilcoxon signed-rank test) is reported in bold. \label{tab:result_tox2}}
\begin{tabular}{|l||c|c|c|c|c|c|c|c|c|c|c|}
\hline
 & \multicolumn{3}{c|}{Blood}& \multicolumn{3}{c|}{Breast} & \multicolumn{3}{c|}{Prostate} &Average\\
 & GI50 & TGI & LC50 & GI50 & TGI & LC50 & GI50 & TGI & LC50 & \\ \hline
%Localized Lasso & 1.051 & 0.632 & 0.516 & 1.118  & 0.606 &  0.594 & 1.369 & 0.492 & 0.521 & {\bf 0.767} \\
Localized Lasso  & 1.030 & 0.622 & 0.529 & 1.129  & 0.627 &  0.562 & 1.297 & 0.518 & 0.539 & {\bf 0.760} \\
% Network Lasso  & 1.111 & 1.053 & 0.956 & 1.325 & 0.821 & 1.086 & 1.596 & 0.730 & 0.690 & 1.041 \\ 
 Network Lasso   & 1.096 & 0.918 & 0.921 & 1.368 & 0.821 & 1.065 & 1.475 & 0.711 & 0.690 & 1.007 \\ 
 FORMULA  & 1.503 & 1.179 & 1.253 & 1.367 & 1.109 & 1.197 & 1.376 & 1.121 & 1.129 & 1.248 \\
 Lasso  & 1.201 & 1.006 & 0.514 & 1.435 & 0.879 & 0.560 & 1.455 & 0.763 & 0.523 & 0.926 \\
Elastic Net  & 1.129 & 0.875 & 0.514 & 1.164 & 0.800 & 0.560 & 1.130 & 0.633 & 0.505 & 0.812 \\
%Kernel Regression  & 1.280 & 0.787 & 0.702 & 1.205 & 0.762 & 0.725 & 1.366 & 0.667 & 0.619 & {\bf 0.901} \\
%Proposed ($\lambda_1 = 1, \lambda_2 = 1$) & 1.47 & 0.97 & 0.51 & 1.45  & 0.82 &  0.55 & 1.46 & 0.72 & 0.53 & 0.94 \\
%Proposed ($\lambda_1 = 10, \lambda_2 = 1$) & 1.46 & 0.96 & 0.51 & 1.42  & 0.76 &  0.56 & 1.45 & 0.73 & 0.54 & 0.93 \\
%Proposed ($\lambda_1 = 10, \lambda_2 = 5$) & 1.49 & 0.98 & 0.51 & 1.46  & 0.82 &  0.56 & 1.46 & 0.73 & 0.54 &  0.95 \\
%Network Lasso ($\lambda_1 = 1$) \cite{hallac2015network} & 1.53 & 1.19 & 1.01 & 1.14 & 0.99 & 0.99 & 1.15 & 0.92 & 0.81 & 1.12 \\ 
%Network Lasso ($\lambda_1 = 10$) \cite{hallac2015network}\! & 1.53 & 1.19 & 1.01 & 1.14 & 0.99 & 0.99 & 1.53 & 0.92 & 0.81 & 1.12\\ 
%Lasso ($\lambda$ = 1)  & 1.38 & 1.28 & 1.51 & 1.43 & 1.01 & 1.09 & 1.66 & 1.05 & 1.09 &1.30 \\
%Lasso ($\lambda$ = 5.0)  & 1.25 & 1.18 & 1.10 & 1.19 & 1.08 & 1.09 & 1.40 & 0.99 & 1.03 & 1.15\\
%Lasso ($\lambda$ = 10.0)  & 1.19 & 1.15 & 1.13 & 1.19 & 1.15 & 1.18 & 1.30 & 1.01 & 1.10 & 1.16\\ \hline 
Kernel Regression  & 1.070 & 0.808 & 0.623 & 1.165 & 0.677 & 0.688 & 1.466 & 0.551 & 0.509 & 0.839 \\
\hline
\end{tabular}
\end{center}}
%\vspace{-.2in}
\end{table*} 

\begin{table*}[t!]
{
\begin{center}
\caption{The number of selected features (genes) on toxicogenomics data. For Localized Lasso, Network Lasso, FORMULA, and Elastic Net, we select features by checking $\|\boldW_{\cdot,i}\|_2 > 10^{-5}$, where $\boldW_{\cdot,i} \in \mathbbR^{n}$ is the $i$-th column of $\boldW$. \label{tab:result_feat}}
\begin{tabular}{|l||c|c|c|c|c|c|c|c|c|c|c|}
\hline
 & \multicolumn{3}{c|}{Blood}& \multicolumn{3}{c|}{Breast} & \multicolumn{3}{c|}{Prostate} &Average\\
 & GI50 & TGI & LC50 & GI50 & TGI & LC50 & GI50 & TGI & LC50 & \\ \hline
% Localized Lasso  & 37.0 & 39.0  & 66.9 & 137.0 & 103.0 &  69.0  &  85.4 & 107.2 & 80.9 & 80.6 \\
Localized Lasso  & 32.7 & 33.4  & 92.5 & 92.6 & 125.9 &  58.2  &  35.1 & 53.9 & 43.4 & 63.4 \\
%Network Lasso \!&  \!1046.7\! & \!1066.8\! & \!1081.5\! & \!1057.1\! & \!1051.5\! & \!1058.2\!  & \!1078.7\! & \!1055.5\!  & \!1053.5\! & \!1051.7\! \\
Network Lasso  \!&  \!1039.6\! & \!1047.3\! & \!1052.2\! & \!1054.6\! & \!1051.5\! & \!1053.3\!  & \!1060.5\! & \!1052.9\!  & \!1053.1\! & \!1061.0\! \\
FORMULA   & 576.6 & 445.6 & 550.5 & 914.3  & 936.7  & 776.3  & 942.0 & 712.2 & 633.6 & 720.8 \\
Lasso & 29.6 & 12.0 & 1.0 & 12.0  & 1.9 &  1.0  & 12.5 & 4.4 & 3.8  & 8.7 \\ %\hline 
Elastic Net & 310.8 & 91.4 & 39.2 & 124.9  & 77.2 &  1.0  & 116.6 & 87.9 & 98.9  & 105.3 \\ %\hline 
%Proposed ($\lambda_1 = 1, \lambda_2 = 1$) & 74.6 & 64.0  & 39.9 & 59.3 & 55.3 &  44.7  &  74.4 & 54.3 & 30.8 & 55.3 \\
%Proposed ($\lambda_1 = 10, \lambda_2 = 1$) & 34.4 & 29.0 & 12.9 & 31.3  & 17.2 & 13.6  & 26.5 & 17.7 & 12.4 & 21.7 \\
%Proposed ($\lambda_1 = 10, \lambda_2 = 5$) & 22.6 & 20.3 & 6.8 & 17.7  & 12.7 & 4.1  & 18.5 & 10.7 & 2.2 & 12.8 \\
%Network Lasso ($\lambda_1 = 1$) \cite{hallac2015network} & \!1095.8\! & \!1096.8\! & \!1100.5\! & \!1101.6\! & \!1101.6\! & \!1103.7\! & \!1099.7\! & \!1099.8\! & \!1100.7 \! & \!1100.0 \! \\
%Network Lasso ($\lambda_1 = 10$) \cite{hallac2015network}\!&  \!1095.4\! & \!1096.7\! & \!1100.5\! & \!1101.5\! & \!1101.5\! & \!1103.7\!  & \!1099.5\! & \!1099.8\!  & \!1100.3\! & \!1099.9\! \\
%Lasso ($\lambda$ = 1)  & 27.2 & 27.3 & 30.65 & 26.55 & 26.35 & 26.75 & 24.55 & 25.75 & 25.55 &  26.74 \\
%Lasso ($\lambda$ = 5.0)   & 6.0 & 6.0 & 5.8 & 3.7  & 7.0  & 6.5  & 7.1 & 4.9 & 5.4 & 5.78 \\
%Lasso ($\lambda$ = 10.0)  & 1.8 & 1.3 & 2.3 & 2.00  & 0.90 &  0.95  & 3.5 & 1.55 & 0.55  & 1.65 \\ %\hline 
%Kernel Regression  & --- & --- & --- & ---  & --- & ---  & --- & --- & --- & --- \\ 
\hline
\end{tabular}
\end{center}}
%\vspace{-.2in}
\end{table*} 

\subsection{Prediction in Toxicogenomics (High-dimensional regression)}
We evaluate our proposed method on the task of predicting toxicity of drugs on three cancer cell lines, based on gene expression measurements. The \emph{Gene Expression} data includes the differential expression of 1106 genes in three different cancer types, for a collection of 53 drugs (i.e., $\boldX_l  = [\boldx_1^{(l)}, \ldots, \boldx_{53}^{(l)}]\in \mathbbR^{1106 \times 53}, l = 1,2,3$). The learning data on \emph{Toxicity} to be predicted contains three dose-dependent toxicity profiles of the corresponding 53 drugs over the three cancers (i.e., $\boldY_l = [\boldy_1^{(l)}, \ldots, \boldy_{53}^{(l)}] \in \mathbbR^{3 \times 53}, l = 1,2,3$). The gene expression data of the three cancers (Blood, Breast and Prostate) comes from the Connectivity Map \cite{lamb2006cmap} and was processed to obtain treatment vs. control differential expression.  The toxicity screening data from the NCI-60 database \cite{shoemaker2006nci60}, summarizes the toxicity of drug treatments in three variables, GI50, LC50 and TGI, representing the 50\% growth inhibition, 50\% lethal concentration, and total growth inhibition levels. The data were confirmed to represent dose-dependent toxicity profiles for the doses used in the corresponding gene expression dataset. %For some of the drugs, the targets are known and similarity of the target information was used as an alternative link graph. A total of 396 known positive interactions of the 53 drugs with 262 targets are known, and hence the graph matrix is quite sparse. However, it is relatively well profiled in comparison to the rest of CMap (2.8\% interactions in 53 drugs, 0.7\% interactions in 881 CMap drugs). Specifically, we first generated the drug-target matrix $\boldS \in \{0,~1\}^{262 \times 53}$, and then computed the graph information as 
%\[
%[\boldR_1]_{ij} =\left\{ \begin{array}{ll}
%\frac{b_{ij}}{b_{ii}} & (i \neq j) \\
%0 & (i = j) \\
%\end{array} \right.,
%\]
%where $\boldB = \boldS^\top \boldS$ and $b_{ij} = [\boldB]_{ij}$. Moreover, 
 
%For some of the drugs, the targets are known and similarity of the target information was used as the graph information. A total of 396 known positive interactions of the 53 drugs with 262 targets are known, and hence the graph matrix is quite sparse. However, it is relatively well profiled in comparison to the rest of CMap (2.8\% interactions in 53 drugs, 0.7\% interactions in 881 CMap drugs). Specifically, we first generated the drug-target matrix $\boldS \in \{0,~1\}^{262 \times 53}$, and then computed the graph information as 
%\[
%r_{ij} =\left\{ \begin{array}{ll}
%\frac{b_{ij}}{b_{ii}} & (i \neq j) \\
%0 & (i = j) \\
%\end{array} \right.,
%\]
%where $\boldB = \boldS^\top \boldS$ and $b_{ij} = [\boldB]_{ij}$.
%\[
%\boldR = (\boldS^\top \boldS)(\text{diag}(\text{diag}^{-1}(\boldS^\top \boldS))^{-1} - \boldI_n,
%\]
%where $\text{diag}^{-1}(\boldS^\top \boldS) \in \mathbbR^{n}$ is the operation to extract the diagonal elements of $\boldS^\top \boldS$. 

In this experiment, we randomly split the data into training and test sets. The training set consisted of 48 drugs and the test set of 5 drugs. Moreover, we introduced a bias term in the proposed method (i.e., $[\boldx^\top~1]^\top \in \mathbbR^{d+1}$), and regularized the entire $\boldw_i$s in the network regularization term, and only $\boldv_i \in \mathbbR^{d-1}$ in the $\ell_{1,2}$ regularization term; here $\boldw_i = [\boldv_i^\top~1]^\top$. We computed the  graph information using the input $\boldX$ as
\begin{align*}
\boldR &= \frac{\boldS^\top + \boldS}{2}, \\
[\boldS]_{ij} &=\left\{ 
\begin{array}{ll} 
1 & \textnormal{$\boldx_j$ is a 5 nearest neighbor of $\boldx_i$ } \\
0 & \textnormal{Otherwise}
\end{array} \right. .
\end{align*}
 We repeated the experiments 20 times and report the average test RMSE scores in Table \ref{tab:result_tox2}. We observed that the proposed localized Lasso outperforms state-of-the-art linear methods. Moreover, the proposed method also outperformed the \emph{nonlinear} kernel regression method, which has high predictive power but cannot identify features. 

 In Table \ref{tab:result_feat}, we report the number of selected features in each method. It is clear that the number of selected features in the proposed method is much smaller than that of the network Lasso. In some cases Lasso and Elastic net selected only one feature. This means that the features were shrunken to zero and only bias term  remained. In summary, the proposed method is suited for producing interpretable sparse models in addition to having high predictive power.

  \begin{figure*}[t!]
  \begin{center}
  \begin{minipage}[t]{0.32\linewidth}
\centering
  {\includegraphics[width=0.99\textwidth]{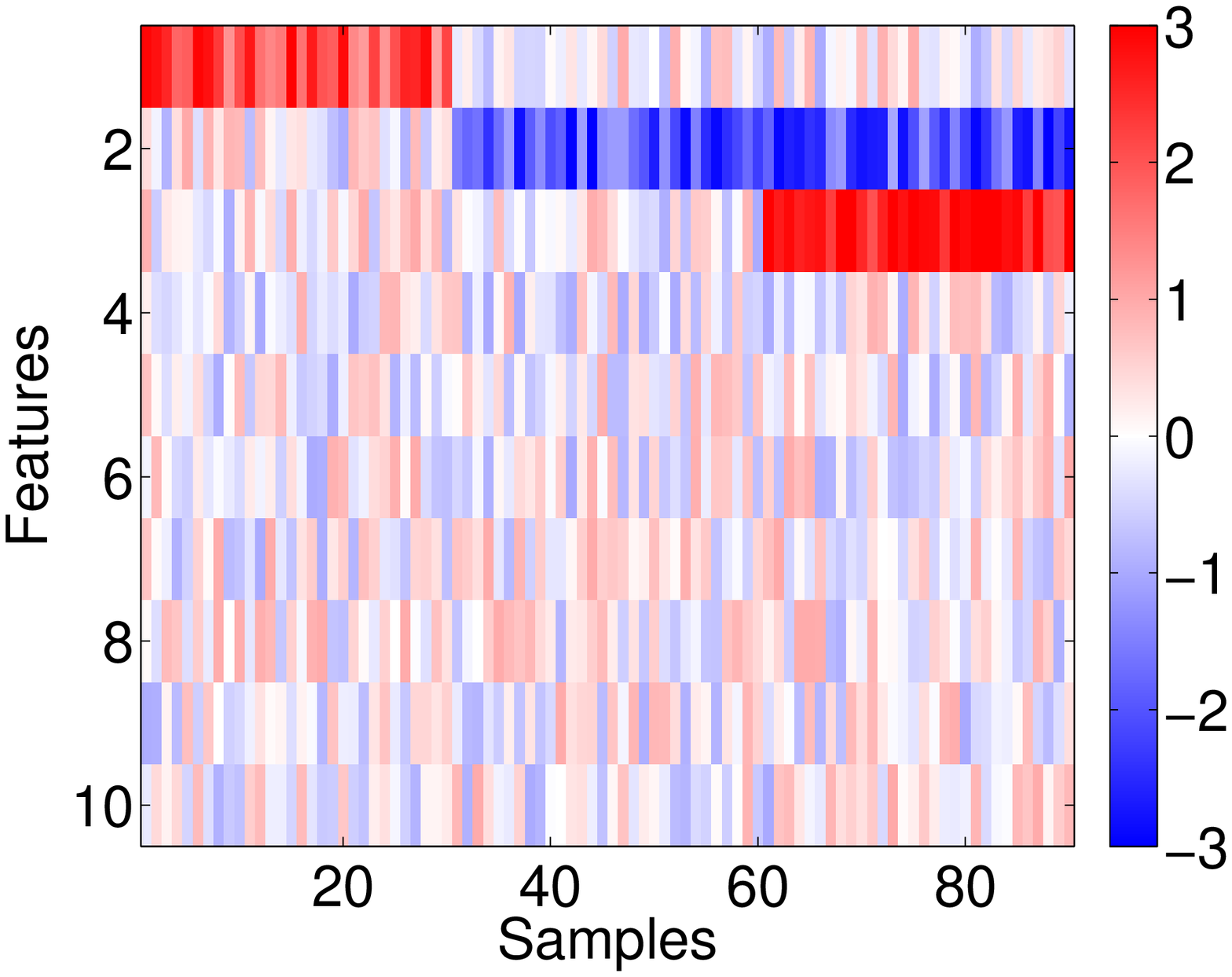}} \\ \vspace{-0.10cm}
(a) Input data  ($\boldX$).
\end{minipage}
\begin{minipage}[t]{0.32\linewidth}
\centering
  {\includegraphics[width=0.99\textwidth]{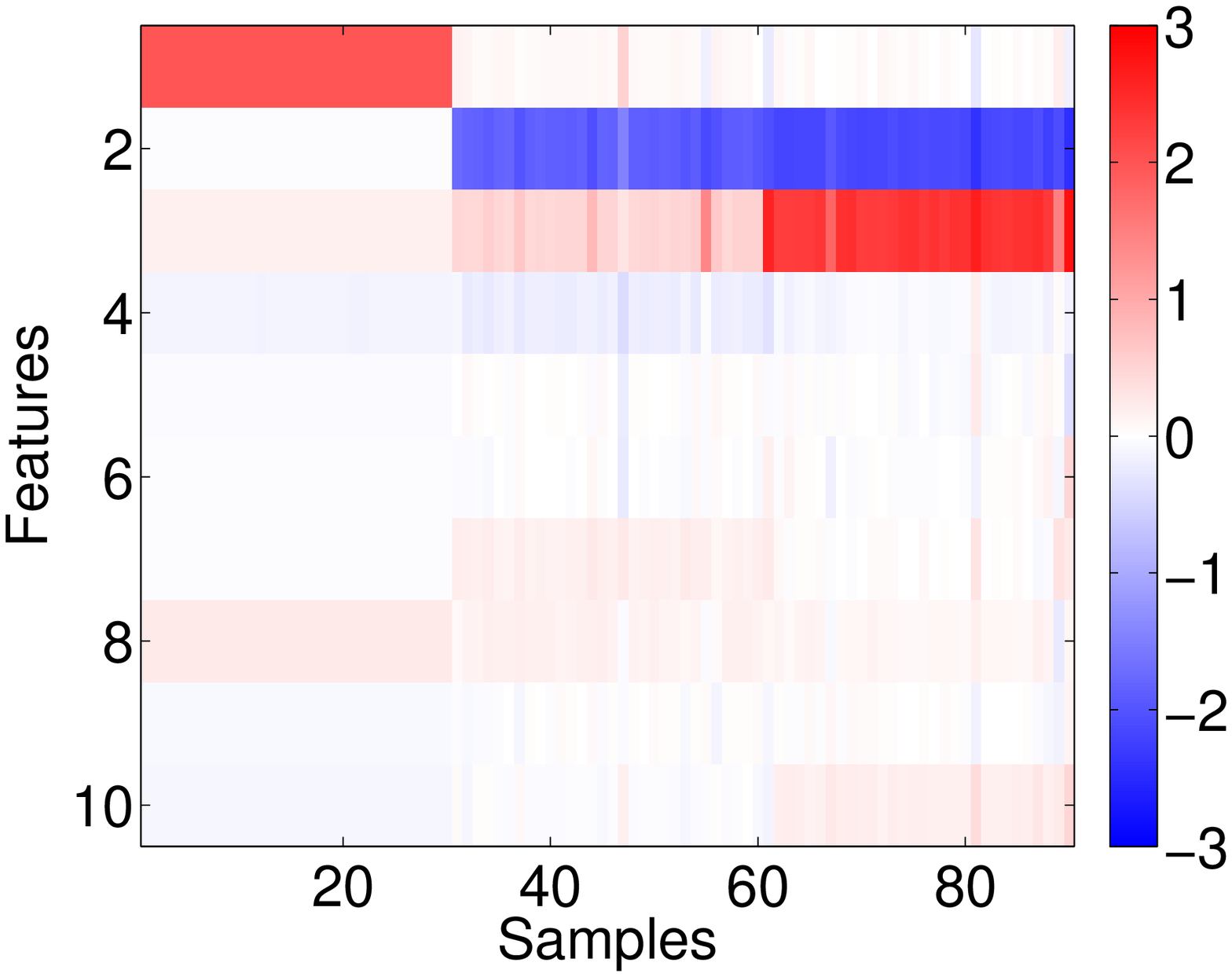}} \\ \vspace{-0.10cm}
(b) Network Lasso.
\end{minipage}
\begin{minipage}[t]{0.325\linewidth}
\centering
  {\includegraphics[width=0.99\textwidth]{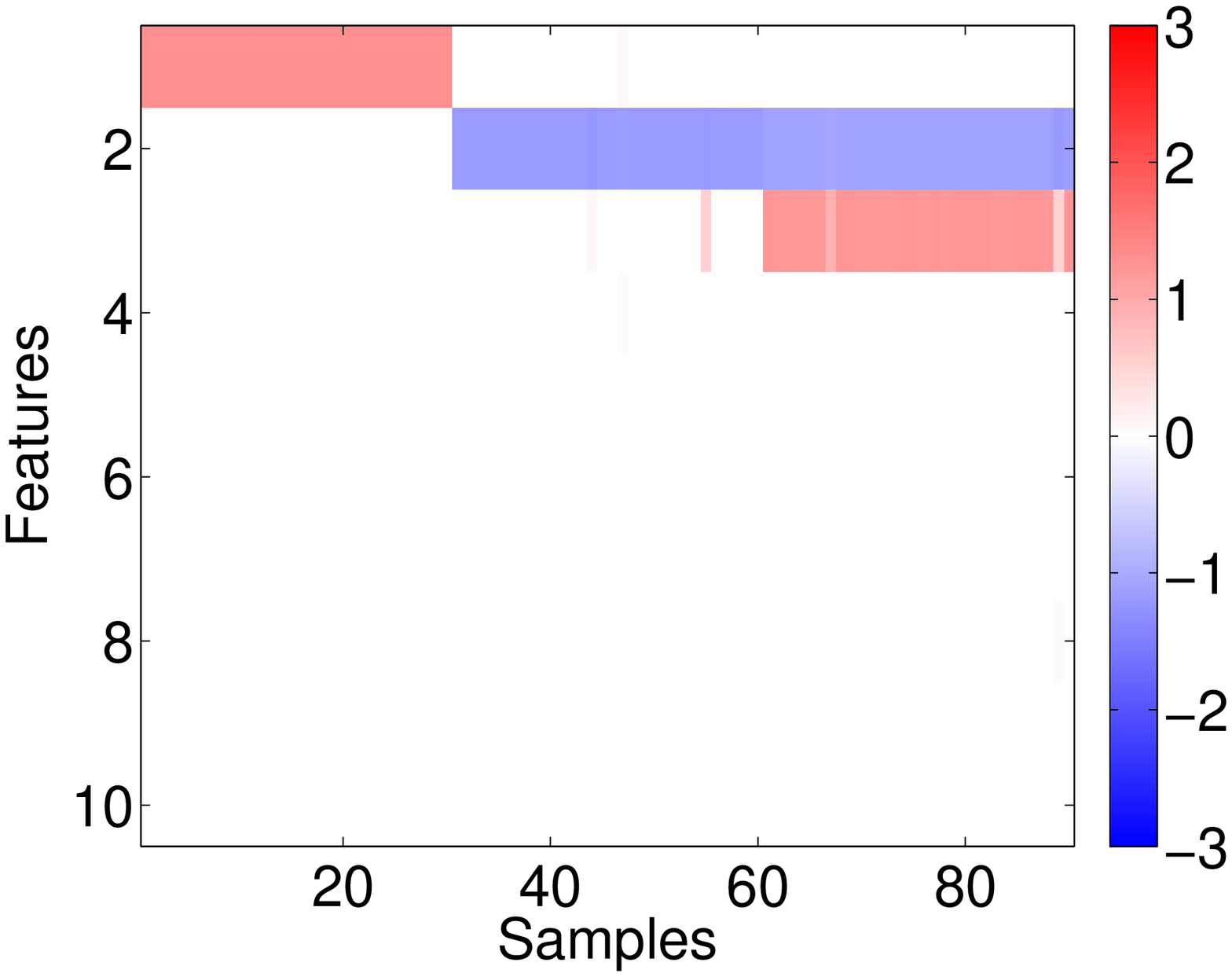}} \\ \vspace{-0.10cm}
(c) Proposed ($\lambda_1 = 5, \lambda_2 = 0.5$).
\end{minipage}
\begin{minipage}[t]{0.325\linewidth}
\centering
  {\includegraphics[width=0.99\textwidth]{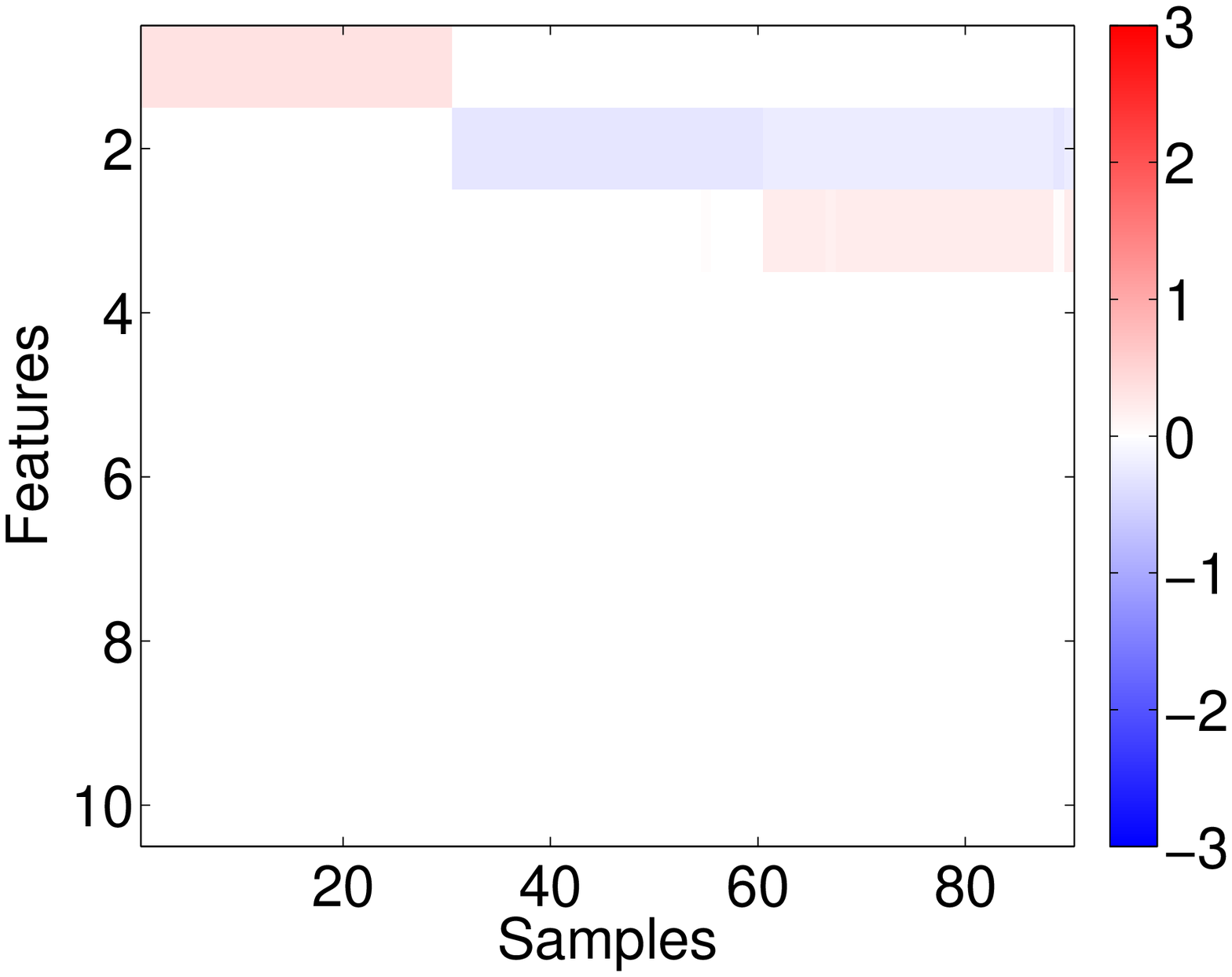}} \\ \vspace{-0.10cm}
(d) Proposed ($\lambda_1 = 5, \lambda_2 = 5$).
\end{minipage}
\begin{minipage}[t]{0.325\linewidth}
\centering
 {\includegraphics[width=0.99\textwidth]{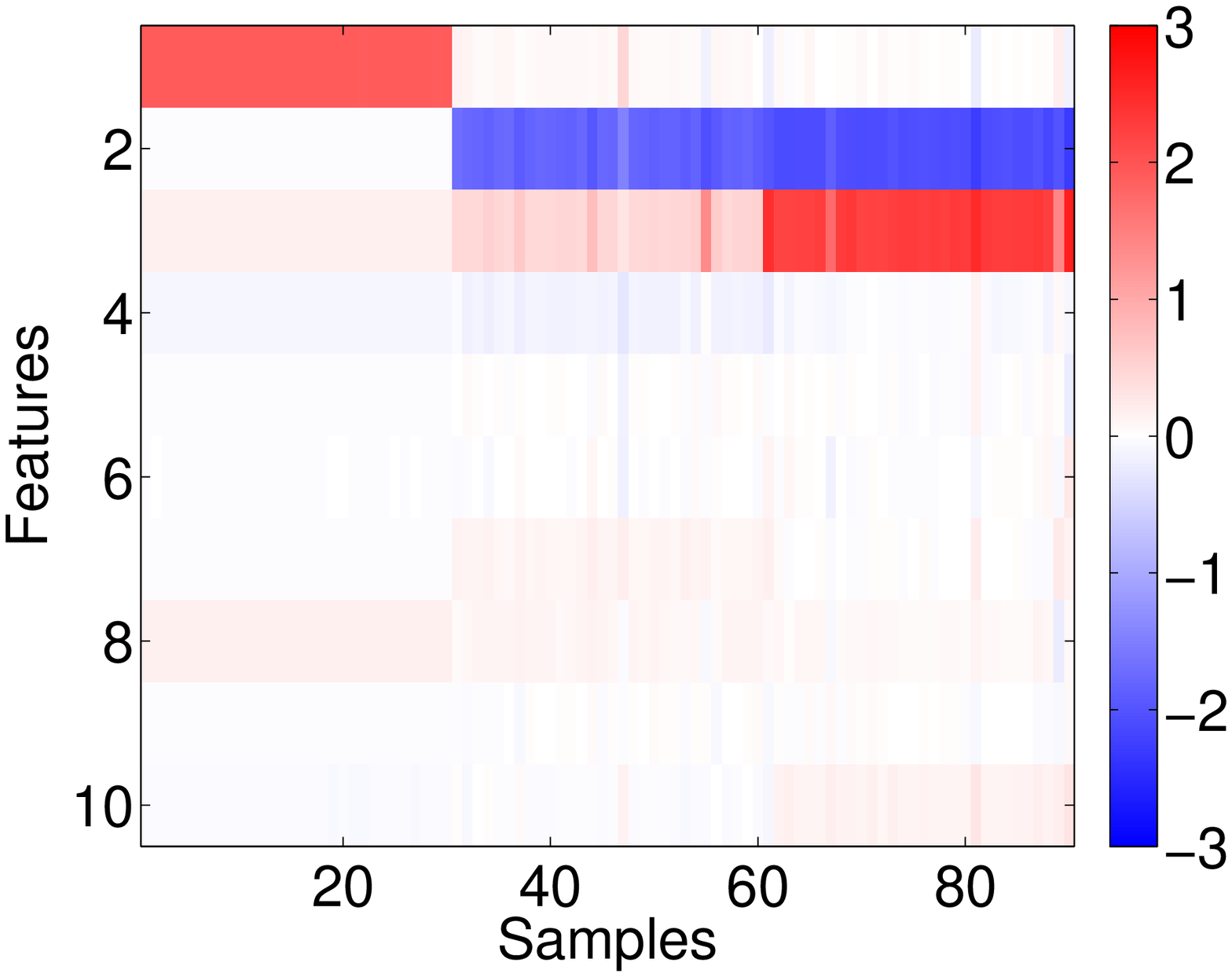}} \\ \vspace{-0.10cm}
(e) Network Lasso + $\ell_{2,1}$ ($\lambda_1 = 5, \lambda_2 = 1$).
\end{minipage} 
\begin{minipage}[t]{0.325\linewidth}
\centering
 {\includegraphics[width=0.99\textwidth]{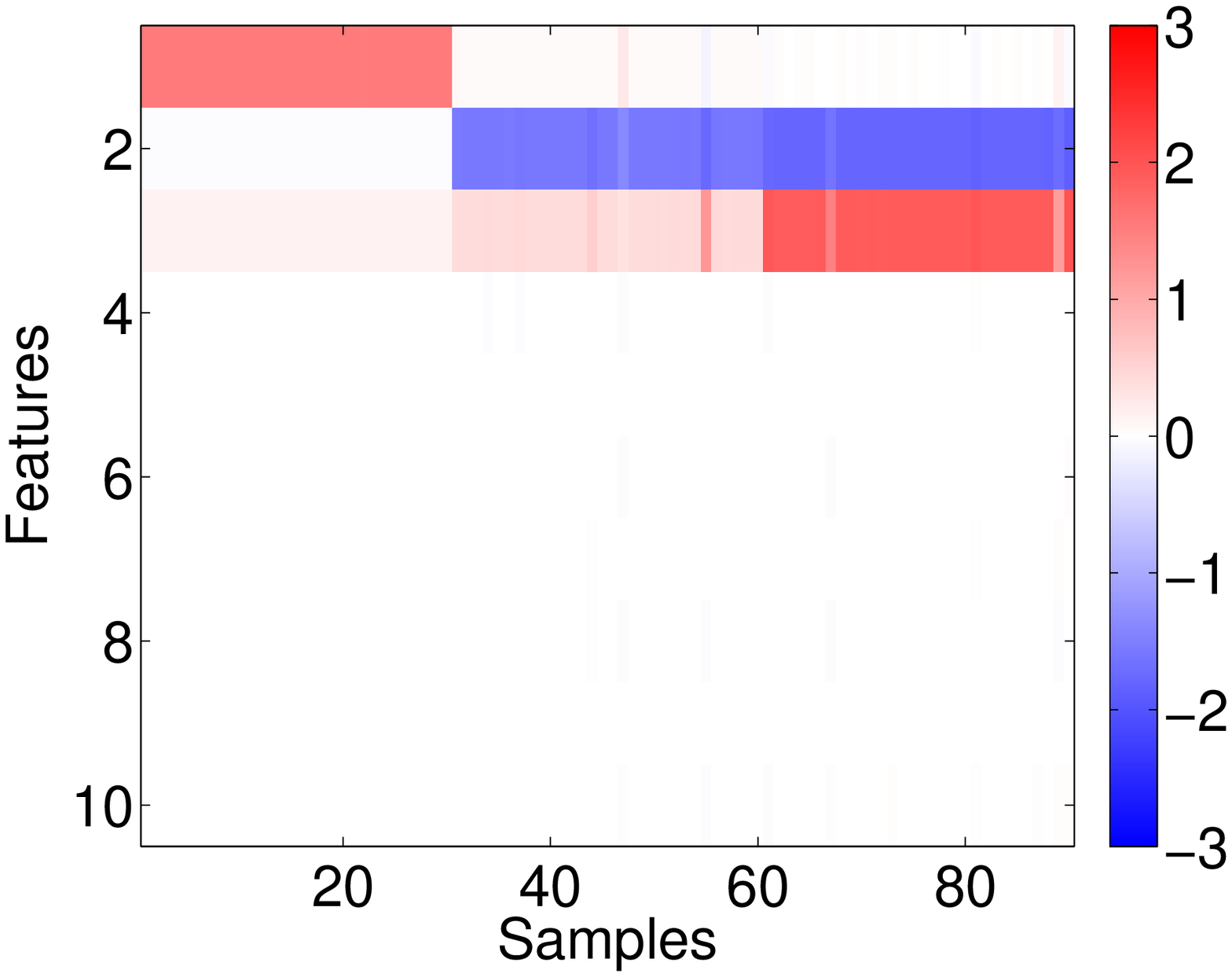}} \\ \vspace{-0.10cm}
(f) Network Lasso + $\ell_{2,1}$ ($\lambda_1 = 5, \lambda_2 = 5$).
\end{minipage}
\caption{The learned coefficient matrix for the synthetic clustering data. For the Network Lasso + $\ell_{2,1}$, \emph{feature-wise} group regularizer is used instead of $\ell_{1,2}$ and $\lambda_2 \geq 0$ is the regularization parameter for the $\ell_{2,1}$ term. (a): The input matrix $\boldX$. (b): Network Lasso ($\lambda_1 = 5, \lambda_2 = 0$). (c): Localized Lasso ($\lambda_1 = 5, \lambda_2 = 0.5$). (d): Localized Lasso ($\lambda_1 = 5, \lambda_2 = 10$). (e): Network Lasso + $\ell_{2,1}$ ($\lambda_1 = 5, \lambda_2 = 1$). (f): Network Lasso + $\ell_{2,1}$ ($\lambda_1 = 5, \lambda_2 = 5$). Note that the Network Lasso + $\ell_{2,1}$ is equivalent to sparse convex clustering \cite{wang2016sparse}. }
    \label{fig:toy_clustering}
\end{center}
\vspace{-.15in}
\end{figure*}

\subsection{Synthetic experiment (Clustering)}
In this section, we illustrate the behavior of the proposed method for convex clustering using a synthetic dataset.

We generated the input variables as
\begin{eqnarray}
x_{ij}\sim \left\{ \begin{array}{ll}
 \text{Unif}(-3,-1) & (i = 1, j = 1, \ldots, 30) \\
 \text{Unif}(1,3) & (i = 2, j = 31, \ldots, 60) \\
 \text{Unif}(2,4) & (i = 3, j = 61,  \ldots 90) \\
 \text{Unif}(-1,1) & \textnormal{Otherwise}\\
\end{array} \right., 
\end{eqnarray} 
where $x_{k,i}$ is the value of the $k$-th feature in the $i$-th sample. In this experiment, we compare our proposed method with the network lasso (i.e., $\lambda_2 = 0$) and  sparse convex clustering \cite{wang2016sparse}, which employs the feature-wise group regularization (i.e., $\ell_{2,1}$-norm) in addition to the network regularization. 

Figure~\ref{fig:toy_clustering} shows the learned coefficient matrices. As can be seen, the weight matrix of the network lasso is non-sparse. In contrast, it is possible to obtain the correct sparsity pattern using the proposed regularization. The sparse convex clustering method \cite{wang2016sparse} can select the correct global set of features, but is less accurate in selecting the local feature sets than the proposed regularization.
 
\begin{table*}[t!]
{
\begin{center}
\caption{Experimental results (ARI) on real-world datasets. Larger ARI is better. $K$ is the number of true clusters. \label{tab:result_clust}}
\begin{tabular}{|l||c|c|c|l|l|l|}
\hline
Data & $d$ & $n$ & $K$ & Localized Lasso & Sparse Conv. Clust.  & Conv. Clust. \\ \hline
%LIBRAS \cite{dias2009hand}  &  90    &   144 &  6 &   0.6661 ($\lambda_1 = 10, \lambda_2 = 10^{-3}$)      &  0.6489 ($\lambda_1 = 9$, $\lambda_2 = 1$)  & 0.6370 ($\lambda_1 = 0$) \\ 
%ORL  \cite{ref:orl}     & 1024   &  400 &   40  &    0.4356 ($\lambda_1=5, \lambda_2=10^{-4}$)    & 0.4203 ($\lambda_1=5, \lambda_2 = 10^{-3}$)  & 0.4203 ($\lambda_1=5$) \\
LUNG     & 3312 & 203   &   5  &   {\bf 0.6316} ($\lambda_1 = 15, \lambda_2 = 1$)  &  0.5692 ($\lambda_1= 9, \lambda_2=8$) & 0.3715 ($\lambda_1 = 10$) \\
COIL20     & 1024   &  1440 &   20  &  {\bf  0.8048} ($\lambda_1=8, \lambda_2=0.1$)    & 0.7795 ($\lambda_1=15, \lambda_2 = 13$)  & 0.6991 ($\lambda_1=15$) \\ 
Lymphoma       & 4026   &  96 &   9  &  {\bf 0.6174} ($\lambda_1=5, \lambda_2=0.01$)    & 0.2673 ($\lambda_1=9, \lambda_2=0$) & 0.2673 ($\lambda_1=9$) \\
\hline
\end{tabular}
\end{center}}
%\vspace{-.2in}
\end{table*} 

\subsection{Benchmark experiments (Clustering)}
We evaluated the proposed sparse convex clustering method on three benchmark datasets. We compared it with the convex clustering \cite{pelckmans2005convex,hocking2011clusterpath} and the sparse convex clustering + $\ell_{2,1}$ \cite{wang2016sparse} algorithms. For all methods, we first ran the clustering algorithm which produced an estimate $\widehat{\boldW}$. Then, we applied an agglomerative clustering algorithm to threshold the $\widehat{\boldW}$ into a disjoint set of cluster indices. The clustering performance was evaluated by the \emph{adjusted Rand index} (ARI) \cite{hubert1985comparing} between the estimated class labels and true labels. We ran each clustering method by multiple regularization parameter values and report the best ARI score. For all methods, the candidate lists of $\lambda_1$ and $\lambda_2$ were $\{0,0.01, 0.1, 1,2,\ldots, 15\}$ and $\{0, 0.01, 0.1, 1,2,\ldots, 15\}$, respectively.

Table~\ref{tab:result_clust} shows the ARI results. As can be seen, the proposed method outperforms the existing state-of-the-art convex clustering methods for high-dimensional clustering problems. In other words, inducing the sample-wise exclusive sparsity is crucial to obtaining better clustering results.

\section{Conclusion}
In this paper, we proposed the localized Lasso method, which can produce sparse interpretable local models for high-dimensional problems. We proposed a simple yet efficient optimization approach by introducing structured sparsity: sample-wise network regularizer and sample-wise exclusive sparsity. Thanks to the structured sparsity, the proposed method had better regression performance with a smaller number of features than the alternatives. Moreover, the sparsity pattern in the learned models aids interpretation. We showed that the proposed method compares favorably with state-of-the-art methods on simulated data and molecular biological personalized medicine data.

%\section*{Acknowledgment}
%We thank Dr. Suleiman A. Khan for providing the toxicity data set. 

\section*{Appendix}
\subsection*{Propositions used for deriving Eq. \eqref{eq:optimum_solution} in main paper}
%We present the propositions and lemma used in main paper.

\begin{prop}
\label{prop_derive_net}
Under $r_{ij} \geq 0$, $r_{ij} = r_{ji}$, $r_{ii} = 0$, we have
\begin{align*}
\frac{\partial}{\partial \textnormal{vec}(\boldW)} \sum_{i,j = 1}^{n} r_{ij} \|{\boldw}_i - {\boldw}_j\|_{2} = 2 \boldF_{g} \textnormal{vec}(\boldW),
\end{align*}
where  
\begin{align*}
\boldF_g &= \boldI_d \otimes \boldC,\\
[\boldC]_{i,j} &=\left\{ \begin{array}{ll}
\sum_{j' = 1}^n \frac{r_{ij'}}{\|\boldw_i - \boldw_j'\|_{2}}  - \frac{r_{ij}}{\|\boldw_i - \boldw_j\|_{2}} & (i = j) \\
\frac{-r_{ij}}{\|\boldw_i - \boldw_j\|_{2}} & (i \neq j) \\
\end{array} \right. .
\end{align*}

\noindent Proof: Under $r_{ij} \geq 0$, $r_{ij} = r_{ji}$, $r_{ii} = 0$, the derivative of the network regularization term with respect to $\boldw_k$ is given as
\begin{align*}
\frac{\partial}{\partial \boldw_k} \sum_{i,j = 1}^{n} r_{ij} \|{\boldw}_i - {\boldw}_j\|_{2} &= \sum_{i = 1}^n r_{ik} \frac{\boldw_k - \boldw_i}{\|\boldw_k - \boldw_i\|_2} + \sum_{j = 1}^n r_{kj} \frac{\boldw_k - \boldw_j}{\|\boldw_j-\boldw_k \|_2}\\
&= \boldw_k \left(\sum_{i = 1}^n \frac{r_{ik}}{\|\boldw_k - \boldw_i\|_2} + \sum_{j = 1}^n \frac{r_{kj}}{\|\boldw_j - \boldw_k\|_2}\right) \\
&\phantom{=} - \sum_{i = 1}^n \frac{r_{ik}}{\|\boldw_k - \boldw_i\|_2} \boldw_i -\sum_{j = 1}^n \frac{r_{kj}}{\|\boldw_j - \boldw_k\|_2} \boldw_j \\
&= 2\left(\boldw_k \sum_{i = 1}^n \frac{r_{ik}}{\|\boldw_k - \boldw_i\|_2}  - \sum_{i = 1}^n \frac{r_{ik}}{\|\boldw_k - \boldw_i\|_2} \boldw_i \right).
\end{align*}
Thus, 
\begin{align*}
\frac{\partial}{\partial \boldW} \sum_{i,j = 1}^{n} r_{ij} \|{\boldw}_i - {\boldw}_j\|_{2} = 2\boldC \boldW,
\end{align*}
where $\boldW = [\boldw_1, \ldots, \boldw_n]^\top \in \mathbbR^{n \times d}$. Since $\textnormal{vec}(\boldC \boldW \boldI_d) = (\boldI_d \otimes \boldC)\textnormal{vec}(\boldW)$, we have
\begin{align*}
\frac{\partial}{\partial \textnormal{vec}(\boldW)} \sum_{i,j = 1}^{n} r_{ij} \|{\boldw}_i - {\boldw}_j\|_{2} = 2  (\boldI_d \otimes \boldC)\textnormal{vec}(\boldW),
\end{align*}
where $\boldI_d \in \mathbbR^{d \times d}$ is the identity matrix and $\textnormal{vec}(\cdot)$ is the vectorization operator.
\proofend
\end{prop}

\begin{prop}
\label{prop_derive_exc}
\begin{align*}
\frac{\partial}{\partial \textnormal{vec}(\boldW)} \sum_{i = 1}^{n}  \|{\boldw}_i\|_{1}^2 = 2\boldF_e \textnormal{vec}(\boldW),
\end{align*}
where  
\begin{align*}
[\boldF_e]_{\ell,\ell} = \sum_{i = 1}^{\ntr} \frac{I_{i,\ell}\|\boldw_{i}\|_1}{[\textnormal{vec}(|\boldW|)]_\ell}.
\end{align*}
Hence, $I_{i,\ell} \in \{0,~1\}$ are the group index indicators: $I_{i,\ell} = 1$ if the $\ell$-th element $[\textnormal{vec}(\boldW)]_{\ell}$ belongs to group $i$ (i.e., $[\textnormal{vec}(\boldW)]_{\ell}$ is the element of $\boldw_i$), otherwise $I_{i,\ell} = 0$.

%\noindent Proof: The derivative of $\|\boldw\|_1^2$ of the $k$-th element of $\boldw$ ($[\boldw]_k$) is given as
%\begin{align*}
%\frac{\partial}{\partial [\boldw]_k} \|\boldw\|_1^2 = 2 \frac{\|\boldw\|_1}{|[\boldw]_k|}[\boldw]_k.
%\end{align*}
%Using the group information of $\boldw_i$s, we can easily derive the derivative. 
\end{prop}

\subsection*{Propositions and lemmas used for deriving Theorem \ref{theo:theo1} in main paper}

\begin{prop}
\label{prop}
Under $r_{ij} \geq 0$, $r_{ij} = r_{ji}$, $r_{ii} = 0$, we have
\begin{align*}
\textnormal{vec}(\boldW)^\top  \boldF_g^{(t)}  \textnormal{vec}(\boldW) &= \sum_{i,j = 1}^{n} r_{ij} \frac{\|{\boldw}_i - {\boldw}_j\|_{2}^2}{2\|{\boldw}_i^{(t)} - {\boldw}_j^{(t)}\|_{2}},
\end{align*}
where 
\begin{align*}
\boldF_g^{(t)} &= \boldI_d \otimes \boldC^{(t)}, \\
[\boldC^{(t)}]_{i,j} &=\left\{ \begin{array}{ll}
\sum_{j' = 1}^n \frac{r_{ij'}}{\|\boldw_{i}^{(t)} - \boldw_{j'}^{(t)}\|_{2}} - \frac{r_{ij}}{\|\boldw_i^{(t)} - \boldw_j^{(t)}\|_{2}} & (i = j) \\
\frac{-r_{ij}}{\|\boldw_i^{(t)} - \boldw_j^{(t)}\|_{2}} & (i \neq j) \\
\end{array} \right. .
\end{align*}
\noindent Proof:
\begin{align*}
&\sum_{i,j = 1}^{n} r_{ij} \frac{\|{\boldw}_i - {\boldw}_j\|_{2}^2}{2\|{\boldw}_i^{(t)} - {\boldw}_j^{(t)}\|_{2}}\\
&= \sum_{i = 1}^n {\boldw}_i^\top {\boldw}_i \sum_{j = 1}^n \frac{r_{ij}}{2\|{\boldw}_i^{(t)} - {\boldw}_j^{(t)}\|_{2}} + \sum_{j = 1}^n {\boldw}_j^\top {\boldw}_j \sum_{i = 1}^n \frac{r_{ij}}{2\|{\boldw}_i^{(t)} - {\boldw}_j^{(t)}\|_{2}} - 2\sum_{i = 1}^n \sum_{j = 1}^n {\boldw}_i^\top {\boldw}_j \frac{r_{ij}}{2\|{\boldw}_i^{(t)} - {\boldw}_j^{(t)}\|_{2}} \\
&= \textnormal{tr}(\boldW^\top \boldC^{(t)} \boldW) \\
&= \textnormal{vec}(\boldW)^\top (\boldI_d \otimes \boldC^{(t)})\textnormal{vec}(\boldW),
\end{align*}
where $\boldI_d \in \mathbbR^{d \times d}$ is the identity matrix, $\textnormal{tr}(\cdot)$ is the trace operator, and $\textnormal{vec}(\cdot)$ is the vectorization operator.

\end{prop}

\begin{lemm}
\label{lemm1}
Under the updating rule of Eq. \eqref{eq:update_beta}, 
\[
\widetilde{J}(\boldW^{(t+1)}) - \widetilde{J}(\boldW^{(t)}) \leq 0.
\]
Proof: Under the updating rule of Eq. \eqref{eq:update_beta}, since Eq.\eqref{eq:objective_function3} is a convex function and the optimal solution is obtained by solving $\frac{\partial \widetilde{J}(\boldW)}{\partial \boldW)} = 0$, the obtained solution $\boldW^{(t+1)}$ is the global solution. That is, $\widetilde{J}(\boldW^{(t+1)}) \leq \widetilde{J}(\boldW^{(t)})$.
\end{lemm}

\begin{lemm}
\label{lemm2}
For any nonzero vectors $\boldw, \boldw^{(t)} \in \mathbbR^{d}$, the following inequality holds \cite{nie2010efficient}:
\[
\|\boldw\|_2 - \frac{\|\boldw\|_2^2}{2\|\boldw^{(t)}\|_2} \leq \|\boldw^{(t)}\|_2 - \frac{\|\boldw^{(t)}\|_2^2}{2\|\boldw^{(t)}\|_2}.
\]
\end{lemm}

\begin{lemm} 
\label{lemm3}
For $r_{i,j} \geq 0, \forall i,j$, the following inequality holds for any non-zero vectors $\boldw_i^{(t)}-\boldw_j^{(t)}, \boldw_{i}^{(t+1)}-\boldw_{j}^{(t+1)}$: %\cite{nie2010efficient}:
\begin{align*}
 &\sum_{i,j = 1}^n r_{ij}\|\boldw_i^{(t+1)} - \boldw_j^{(t+1)}\|_2 - \textnormal{vec}(\boldW^{(t+1)})^\top \boldF_g^{(t)} \textnormal{vec}(\boldW^{(t+1)}) \nonumber \\
 &- \left(\sum_{i,j = 1}^n r_{ij}\|\boldw_i^{(t)} - \boldw_j^{(t)}\|_2 - \textnormal{vec}(\boldW^{(t)})^\top \boldF_g^{(t)} \textnormal{vec}(\boldW^{(t)})\right) \leq 0.
 %\nonumber \\
%& \leq \sum_{j = 1}^d \|\boldw_j^{(t)}\|_2^2 - \text{vec}(\boldW^{(t)})^\top \boldF_g^{(t)} \text{vec}(\boldW^{(t)}) 
%\sum_{j = 1}^d \|\boldw_j^{(t+1)}\|_2^2 - \sum_{j = 1}^d \frac{\|\boldw_{j}^{(t+1)}\|_2^2}{2\|\boldw_j^{(t)}\|_2^2} \leq \sum_{j = 1}^d \|\boldw_j^{(t)}\|_2^2 - \sum_{j = 1}^d \frac{\|\boldw_{j}^{(t)}\|_2^2}{2\|\boldw_j^{(t)}\|_2^2}
\end{align*}

\noindent Proof: $ \textnormal{vec}(\boldW)^\top \boldF_g^{(t)} \textnormal{vec}(\boldW)$ can be written as
\begin{align*}
&\textnormal{vec}(\boldW)^\top \boldF_g^{(t)} \textnormal{vec}(\boldW) = \sum_{i,j = 1}^n r_{ij}\frac{\|\boldw_i - \boldw_j\|_2^2}{2\|\boldw_i^{(t)} - \boldw_j^{(t)}\|_2}.
\end{align*}
where $r_{ij} \geq 0$.

Then, the left hand side  equation can be written as
\begin{align*}
\Delta_g &= \sum_{i,j = 1}^n r_{ij} \left( \|\boldw_i^{(t+1)} - \boldw_j^{(t+1)}\|_2 - \frac{\|\boldw_i^{(t+1)} - \boldw_j^{(t+1)}\|_2^2}{2\|\boldw_i^{(t)} - \boldw_j^{(t)}\|_2}\right)\\
&\phantom{=}- \sum_{i,j = 1}^n r_{ij}\left( \|\boldw_i^{(t)} - \boldw_j^{(t)}\|_2 -  \frac{\|\boldw_i^{(t)} - \boldw_j^{(t)}\|_2^2}{2\|\boldw_i^{(t)} - \boldw_j^{(t)}\|_2} \right).
\end{align*}
Using Lemma \ref{lemm2}, $\Delta_g \leq 0$.\proofend
\
\end{lemm}

%\begin{prop}
%$\sum_{k = 1}^n \|\widetilde{\boldw}_{k}\|_1^2 =  \textnormal{vec}(\boldW)^\top \boldF_e \textnormal{vec}(\boldW)$
%\end{prop}

\begin{lemm} 
\label{lemm4}
The following inequality holds for any non-zero vectors \cite{kong2014exclusive}:
\begin{align}
& \sum_{i = 1}^{\ntr} \|{\boldw}_{i}^{(t+1)}\|_1^2 -  \textnormal{vec}(\boldW^{(t+1)})^\top \boldF_e^{(t)} \textnormal{vec}(\boldW^{(t+1)}) \nonumber \\
&- \left(\sum_{i = 1}^{\ntr} \|{\boldw}_{i}^{(t)}\|_1^2 -  \textnormal{vec}(\boldW^{(t)})^\top \boldF_e^{(t)} \textnormal{vec}(\boldW^{(t)})\right) \leq 0. %\sum_{\ell = 1}^{dn} \sum_{k = 1}^n \frac{I_{k,\ell}\|\widetilde{\boldw}_{k}^{(t)}\|_1}{|[\text{vec}(\boldW^{(t)}]_\ell|} ([\text{vec}(\boldW^{(t+1)}]_\ell)^2 %\text{vec}(\boldW^{(t)})^\top \boldF_e^{(t)} \text{vec}(\boldW^{(t)})
\end{align}

\noindent Proof: $ \textnormal{vec}(\boldW)^\top \boldF_e^{(t)} \textnormal{vec}(\boldW)$ can be written as
\begin{align*}
\textnormal{vec}(\boldW)^\top \boldF_e^{(t)} \textnormal{vec}(\boldW) &= \sum_{\ell = 1}^{d\ntr} [\textnormal{vec}(\boldW)]_\ell^2 \sum_{i = 1}^{\ntr} \frac{{I}_{i,\ell}\|{\boldw}_{i}^{(t)}\|_1}{[\textnormal{vec}(|\boldW^{(t)}|)]_\ell}  \\
&= \sum_{i = 1}^{\ntr}  \left(\sum_{j = 1}^d \frac{[{\boldw}_i]_j^2}{[|{\boldw}_i^{(t)}|]_j} \right)\|{\boldw}_{i}^{(t)}\|_1. 
\end{align*}
Thus, the left hand equation is written as
\begin{align*}
\Delta_e %&= \sum_{k = 1}^{\ntr} \|\widetilde{\boldw}_{k}^{(t+1)}\|_1^2 -   \sum_{k = 1}^{\ntr} \left(\sum_{j = 1}^d \frac{[\widetilde{\boldw}_k^{(t+1)}]_j^2}{[|\widetilde{\boldw}_k^{(t)}|]_j} \right) \|\widetilde{\boldw}_{k}^{(t)}\|_1 , \\
%& = \sum_{k = 1}^{\ntr} \|\widetilde{\boldw}_{k}^{(t+1)}\|_1^2 -   \left(\sum_{j = 1}^d \frac{[\widetilde{\boldw}_k^{(t+1)}]_j^2}{[|\widetilde{\boldw}_k^{(t)}|]_j} \right) \|\widetilde{\boldw}_{k}^{(t)}\|_1\\
&= \sum_{i = 1}^{\ntr}\! \left[\! \left(\!\sum_{j = 1}^d [{\boldw}_{i}^{(t+1)}]_j\! \right)^2\! -\!  \left(\!\sum_{j = 1}^d \frac{[{\boldw}_i^{(t+1)}]_j^2}{[|{\boldw}_i^{(t)}|]_j} \right)\!\!\left( \!\sum_{j = 1}^d [{\boldw}_{i}^{(t)}]_j \!\right) \!\right] \\
&= \sum_{i = 1}^{\ntr} \!\left[ \!\left(\!\sum_{j = 1}^d a_j^{(t)} b_j^{(t)}\! \right)^2\! -\! \left(\!\sum_{j = 1}^d {(a_j^{(t)})}^2 \! \right)\!\! \left(\!\sum_{j = 1}^d\! (b_j^{(t)})^2\!\right)\!\right] \leq 0,
\end{align*}
where $a_j^{(t)} = \frac{[|{\boldw}_i^{(t+1)}|]_j}{\sqrt{[|{\boldw}_i^{(t)}|]_j}}$ and $b_j^{(t)} =\sqrt{[|{\boldw}_i^{(t)}|]_j} $, and $ \textnormal{vec}(\boldW^{(t)})^\top \boldF_e^{(t)} \textnormal{vec}(\boldW^{(t)}) =  \sum_{i = 1}^{\ntr} \|{\boldw}_{i}^{(t)}\|_1^2$. The inequality holds due to cauchy inequality \cite{steele2004introduction}. \proofend
\end{lemm}

\begin{lemm}
\label{lemm5}
Under the updating rule of Eq. \eqref{eq:update_beta}, 
\[
{J}(\boldW^{(t+1)}) - {J}(\boldW^{(t)}) \leq \widetilde{J}(\boldW^{(t+1)}) - \widetilde{J}(\boldW^{(t)}).
\]
\noindent Proof: The difference between the right and left side equations is given as  
\begin{align*}
\Delta& = {J}(\boldW^{(t+1)}) - {J}(\boldW^{(t)}) - (\widetilde{J}(\boldW^{(t+1)}) - \widetilde{J}(\boldW^{(t)})) \\
&= \lambda_1 {\huge \textnormal{(}}\sum_{i,j = 1}^n r_{ij}\|\boldw_i^{(t+1)} - \boldw_j^{(t+1)}\|_2 - \textnormal{vec}(\boldW^{(t+1)})^\top \boldF_g^{(t)} \textnormal{vec}(\boldW^{(t+1)}) \\
&\phantom{=} -{\Large \textnormal{[}}\sum_{i,j = 1}^n r_{ij}\|\boldw_i^{(t)} - \boldw_j^{(t)}\|_2 - \textnormal{vec}(\boldW^{(t)})^\top \boldF_g^{(t)} \textnormal{vec}(\boldW^{(t)}){\Large \textnormal{]}}{\huge \textnormal{)}}\\
&\phantom{=} + \lambda_2 {\huge \textnormal{(}}\sum_{i = 1}^{\ntr} \|{\boldw}_{i}^{(t+1)}\|_1^2 -  \textnormal{vec}(\boldW^{(t+1)})^\top \boldF_e^{(t)} \textnormal{vec}(\boldW^{(t+1)}) \\
&\phantom{=} - {\Large \textnormal{[}}\sum_{i = 1}^{\ntr} \|{\boldw}_{i}^{(t)}\|_1^2 -  \textnormal{vec}(\boldW^{(t)})^\top \boldF_e^{(t)} \textnormal{vec}(\boldW^{(t)}){\Large \textnormal{]}}{\huge \textnormal{)}} 
\end{align*}
\end{lemm}
Based on Lemma \ref{lemm3} and \ref{lemm4}, $\Delta \leq 0$. \proofend %This completes the proof.

{%\small
\bibliography{main}
\bibliographystyle{unsrt}
}

\end{document}

%% file: mathdef.tex
\newtheorem{defi}{Definition}
\newtheorem{prop}[defi]{Proposition}
\newtheorem{lemm}[defi]{Lemma}
\newtheorem{theo}[defi]{Theorem}

\newcommand{\proofend}{\hfill$\Box$\vspace{2mm}}

\newcommand{\diag}[1]{\mathrm{diag}\left(#1\right)}

\newcommand{\mathbbR}{\mathbb{R}}

\newcommand{\boldzero}{{\boldsymbol{0}}}
\newcommand{\boldone}{{\boldsymbol{1}}}

\newcommand{\boldC}{{\boldsymbol{C}}}

\newcommand{\boldF}{{\boldsymbol{F}}}

\newcommand{\boldH}{{\boldsymbol{H}}}
\newcommand{\boldI}{{\boldsymbol{I}}}

\newcommand{\boldR}{{\boldsymbol{R}}}
\newcommand{\boldS}{{\boldsymbol{S}}}

\newcommand{\boldW}{{\boldsymbol{W}}}
\newcommand{\boldX}{{\boldsymbol{X}}}
\newcommand{\boldY}{{\boldsymbol{Y}}}
\newcommand{\boldZ}{{\boldsymbol{Z}}}

\newcommand{\boldf}{{\boldsymbol{f}}}

\newcommand{\boldr}{{\boldsymbol{r}}}

\newcommand{\boldu}{{\boldsymbol{u}}}
\newcommand{\boldv}{{\boldsymbol{v}}}
\newcommand{\boldw}{{\boldsymbol{w}}}
\newcommand{\boldx}{{\boldsymbol{x}}}
\newcommand{\boldy}{{\boldsymbol{y}}}

\newcommand{\ntr}{n}